\documentclass{article}
\usepackage{ijcai25}

\usepackage{times}
\usepackage{soul}
\usepackage{url}
\usepackage[hidelinks]{hyperref}
\usepackage[utf8]{inputenc}
\usepackage[small]{caption}
\usepackage{graphicx}
\usepackage{amsmath}
\usepackage{amsthm}
\usepackage{booktabs}
\usepackage{algorithm}
\usepackage{algorithmic}
\usepackage[switch]{lineno}
\usepackage{url}            
\usepackage{amsfonts}       
\usepackage{nicefrac}       
\usepackage{microtype}      
\usepackage{lipsum}
\usepackage{graphicx}
\usepackage{float}
\usepackage{orcidlink}
\usepackage{multirow}
\graphicspath{ {./images/} }
\usepackage{amsmath}
\usepackage{glossaries}   
\usepackage{hyperref}
    
\title{CognArtive: Large Language Models for Automating Art Analysis and Decoding Aesthetic Elements}

\author{
Afshin Khadangi\orcidlink{0000-0002-0496-5219}$^1$
\and
Amir Sartipi\orcidlink{0000-0002-0124-9823}$^1$\and
Igor Tchappi\orcidlink{0000-0001-5437-1817}$^1$\and
Gilbert Fridgen\orcidlink{0000-0001-7037-4807}$^{1}$\\
\affiliations
$^1$SnT, University of Luxembourg\\
\emails
\{afshin.khadanki, amir.sartipi, igor.tchappi, gilbert.fridgen\}@uni.lu
}

\newacronym{ai}{AI}{Artificial Intelligence}
\newacronym{llm}{LLM}{Large Language Model}
\newacronym{mllm}{MLLM}{Multimodal Large Language Model}
\newacronym{nlp}{NLP}{natural language processing}
\newacronym{ml}{ML}{machine learning}
\newacronym{dl}{DL}{Deep learning}

\setglossarystyle{list} 
\def\BibTeX{{\rm B\kern-.05em{\sc i\kern-.025em b}\kern-.08em
    T\kern-.1667em\lower.7ex\hbox{E}\kern-.125emX}}

\setglossarystyle{list} 

\begin{document}

\maketitle

\begin{abstract}
Art, as a universal language, can be interpreted in diverse ways, with artworks embodying profound meanings and nuances. The advent of \glspl{llm} and the availability of \glspl{mllm} raise the question of how these transformative models can be used to assess and interpret the artistic elements of artworks. While research has been conducted in this domain, to the best of our knowledge, a deep and detailed understanding of the technical and expressive features of artworks using \glspl{llm} has not been explored. In this study, we investigate the automation of a formal art analysis framework to analyze a high-throughput number of artworks rapidly and examine how their patterns evolve over time. We explore how \glspl{llm} can decode artistic expressions, visual elements, composition, and techniques, revealing emerging patterns that develop across periods. Finally, we discuss the strengths and limitations of \glspl{llm} in this context, emphasizing their ability to process vast quantities of art-related data and generate insightful interpretations. Due to the exhaustive and granular nature of the results, we have developed interactive data visualizations, available online~\href{https://cognartive.github.io/}{\textcolor{blue}{https://cognartive.github.io/}}, to enhance understanding and accessibility.
\end{abstract}

\section{Introduction}
Art serves as a universal language, capable of conveying multiple meanings and offering diverse interpretations. Artists, through their unique expressions, provide fresh insights into various concepts, influencing us on both personal and global levels. While traditional analyses of art focus on elements like color, line, and shape, and principles such as balance, contrast, and proportion, a more profound understanding emerges when considering factors beyond the physical. Artists deliberate on materials, scale, composition, and techniques, often infusing their work with emotion, reflecting the zeitgeist, or utilizing light to manipulate mood and meaning~\cite{hodge2024elements}. We can depart from conventional analyses~\cite{10.24963/ijcai.2024/849} by exploring elements that delve deeper into the essence of artworks, aiming to make the understanding of art more compelling and comprehensive. Such framework encourages a richer, more stimulating engagement with art by examining the reasons behind its creation and the diverse approaches of artists, revealing dynamic methods of questioning and interpreting art~\cite{hodge2024elements}.

\gls{ai} in general and \glspl{llm} in particular have emerged as powerful tools with the potential to revolutionize art analysis~\cite{akhtar2024unveiling}. \glspl{llm} are trained on massive datasets of text, enabling them to perform tasks such as translation, summarization, and question answering~\cite{kumar2024large}. By leveraging computer vision, \gls{ml}, and \gls{nlp}, we can extract meaningful information from digital images of art~\cite{castellano2022leveraging,castellano2022deep,cetinic2019deep,imran2023artistic,ijcai2024p0865}. These techniques enable computers to detect visual elements, classify styles, identify artists, and even generate descriptions of artworks~\cite{bin2024gallerygpt}. 

The advent of \glspl{llm} with visual capabilities, often referred to as vision \glspl{llm}, offers a transformative approach to art analysis, expanding the scope of inquiry beyond traditional methods. These models, such as GPT-4V\footnote{\url{https://openai.com/index/gpt-4v-system-card/}}, Flamingo~\cite{alayrac2022flamingo}, and LLaVA~\cite{liu2024visual}, are capable of processing and interpreting visual data in conjunction with textual information, enabling them to identify artistic styles, recognize artists, describe scene elements, and even infer emotional content within artworks. By leveraging vast datasets of images and text, vision \glspl{llm} can provide insights into the formal elements, contextual factors, and cultural significance of art pieces, assisting in tasks like attribution, stylistic classification, analysis of composition, color, and other visual elements~\cite{bin2024gallerygpt,cao2024optimizing,constantinides2024culturai,ozaki2024towards,hayashi2024towards,tao2024does}. While still in their early stages of development, these models demonstrate immense potential to enhance our understanding of art by automating analytical processes and uncovering hidden connections between different art pieces and their history, offering a novel lens through which to appreciate and examine the multifaceted world of artistic creation.

GPT-4V represents a significant leap forward in \gls{ai} technology, enabling a new level of interaction between humans and machines. By incorporating visual input, GPT-4V expands the possibilities of \gls{ai} applications and opens up new avenues for innovation. As this multimodal technology continues to evolve, it holds immense potential to transform the way we interpret visual aspects of the media such as artworks. Moreover, by leveraging systems like Gemini 2.0\footnote{\url{https://deepmind.google/technologies/gemini/}} in conjunction with GPT-4V, we can automate and streamline art analysis in an unprecedented fashion to unveil new insights into the trajectory of art evolution throughout various historical periods.

To this end, we propose to utilize GPT-4V together with Gemini 2.0 and GPT-4 to automate a formal, technical and comprehensive art analysis framework~\cite{hodge2024elements} by analyzing more than 15,000 artworks from 23 prominent artists of all time. To the best of our knowledge, this study presents one of the first approaches to streamline high-throughput and large-scale art analysis by leveraging \glspl{mllm} to assess and interpret artistic elements, providing insights from an expert framework perspective. By automating the process of formal art critique, our study offers a more efficient and objective method for analyzing the technical and aesthetic elements of art, leveraging both quantitative and qualitative metrics.

The remainder of this paper is structured as follows: Section 2 reviews the related works of research on the application of AI and \glspl{llm} in the context of the art analysis. Section 3 describes the methodology used in this study followed by Section 4 that presents our results. Section 5 offers the evaluation of our results, and Section 6 provides a concluding discussion of these results and outlines potential directions for future research.

\section{Related Works}
The study of computational aesthetics and art analysis has grown significantly with advancements in \gls{ai}\gls{ml}. These technologies have provided new methods for exploring artistic styles and understanding cultural patterns. Datasets play an essential role in this progress, serving as the foundation for training and testing models. The quality and diversity of datasets greatly impact model performance and generalizability to various artistic and cultural contexts. Over recent decades, many initiatives have contributed to digitizing art collections, making them accessible for computational analysis and enabling the creation of robust datasets.

One of the most well-known datasets for aesthetic evaluation is the aesthetic visual analysis (AVA) dataset, created by~\cite{murray2012ava}. It contains over 250,000 images, each annotated with scores that reflect their aesthetic quality. This dataset was primarily focused on photographs, as was the aesthetic and attribute database (AADB) introduced by~\cite{kong2016photo} containing aesthetic scores and meaningful attributes assigned to each image by multiple human raters.
 While these datasets helped researchers study the aesthetics of photographs, they were not suitable for analyzing artistic works such as paintings and drawings. To address this, art-specific datasets like WikiArt\footnote{\url{https://www.wikiart.org/}} and the Rijksmuseum Challenge~\cite{mensink14icmr} introduced thousands of images of artworks, enabling tasks such as artist identification and style recognition. 

\gls{dl} methods have advanced the analysis of art, making it possible to classify styles \cite{SCARINGI2025112857} and attribute artworks to their creators with high accuracy. One common approach is transfer learning, where convolutional neural networks trained on large datasets like ImageNet~\cite{deng2009imagenet} are fine-tuned for tasks specific to art analysis~\cite{cetinic2022understanding}. This method was effective for style classification and artist attribution. Neural style transfer (NST)~\cite{jing2019neural,gatys2015neural} is another approach allowing models to separate an artwork's style from its content. NST contributed for further research into how stylistic features can be analyzed and even replicated computationally. Researchers have also studied how styles evolve and connect with cultural patterns. For example,~\cite{cetinic2022understanding} showed that neural networks could detect stylistic features that align with historical concepts in art, while datasets like OmniArt~\cite{strezoski2018omniart} have been used to map stylistic diversity across different cultures. 

However, while \gls{dl} models have significantly transformed the analysis of art by identifying, for example, patterns and predict aesthetic scores, they often struggle to interpret the symbolic meanings or historical significance of artworks~\cite{cetinic2022understanding}. Moreover, \gls{dl} models rely heavily on labeled dataset while creating such datasets for art analysis is time-consuming and expensive knowing that they often focus on a limited range of artistic styles and concepts. Furthermore, models designed for one type of artistic task, such as style classification, struggle to adapt to other tasks without significant adjustments~\cite{cetinic2022understanding,elgammal2018shape}.
\begin{figure*}[!ht] 
    \centering
    \includegraphics[width=\textwidth]{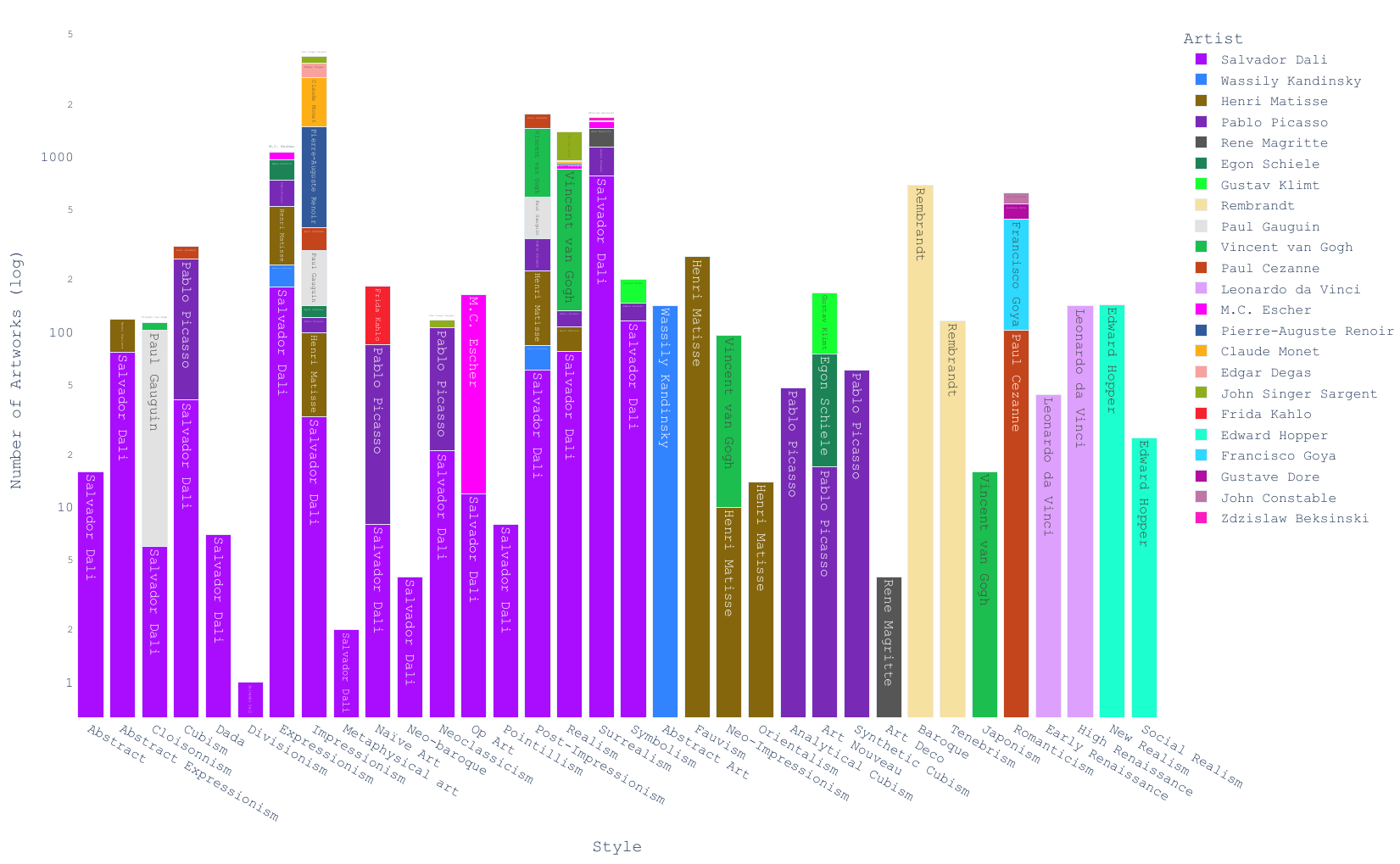}
    \caption{Distribution of the number of artworks across different styles for individual artists. We retrieved more than 15,000 artworks across 23 artists for our study.}
    \label{fig:fig1}
\end{figure*}
Multimodal approaches, which combine text and images, have become an important area of research. Projects like Artpedia~\cite{stefanini2019artpedia} and SemArt~\cite{garcia2018read} introduced models that align both visual and textual description (contextual descriptions)  in a shared semantic space. 
This alignment enables advanced searches based on content. 
For example, a user could search for paintings that match a specific theme or historical context, and the system would retrieve relevant results by analyzing both visual features and textual descriptions. The OmniArt dataset has been particularly useful in this field, as it includes detailed metadata alongside visual and textual annotations, supporting deeper studies of iconography and stylistic elements. These methods highlight the potential of combining different types of data to improve how we study and interpret art.

The rise of LLM such as GPT~\cite{brown2020language},  Gemeni~\cite{team2023gemini}, and CLIP~\cite{radford2021learning} has further contributed to improve computational aesthetics by bridging the gap between visual and textual data. \glspl{llm} have expanded the possibilities for multimodal analysis, allowing models to process text and images in a unified framework. For example, CLIP aligns textual and visual embeddings, enabling the evaluation of artworks based on textual descriptions, such as ``a serene landscape with balanced composition''. Such models offer new capabilities for integrating narrative and aesthetic elements, facilitating tasks like multimodal retrieval and aesthetic judgment.  Generative \glspl{llm}\cite{reddy2021dall,10380595} generates artworks based on textual prompts, providing insights into aesthetic preferences and stylistic variations.

\glspl{llm} with their advanced ability to process and understand complex textual data, can provide an interesting approach to interpreting and analyzing artworks. In fact, \glspl{llm} can analyze not just the visual elements of an artwork but also its associated textual descriptions, historical background, and cultural significance.  This enables them to bridge the gap between visual content and human interpretation, offering deeper insights into the meaning, symbolism, and emotional resonance of art. Glimpses of \glspl{llm} potential in art analysis have emerged, yet a framework that comprehensively examines both technical and conceptual elements of art remains underexplored, motivating the need for this study.

\section{Methods}
\subsection{Dataset}
We retrieved dataset for our study by obtaining a collection of digital reproductions of artworks sourced from WikiArt, a comprehensive online repository of visual art. We focused on a selection of 23 artists widely regarded as among the most influential figures in art history. The dataset encompasses a total of 15,000 artworks across 34 different ground-truth art styles, spanning a broad chronological range from the 15th to the 21st centuries (approximately the 1400s to the 2000s).

As shown in Figure \ref{fig:fig1}, the artists included in our study were selected based on their historical significance, impact on art movements, and enduring popularity. This selection includes well-known artists such as Pablo Picasso, renowned for his pioneering contributions to Cubism; Vincent van Gogh, known for his emotionally charged Post-Impressionist works; Claude Monet, a leading figure of Impressionism; Leonardo da Vinci, the quintessential Renaissance polymath; and Salvador Dalí, a key figure in Surrealism.
\begin{figure*}[!ht] 
    \centering
    \includegraphics[width=\textwidth]{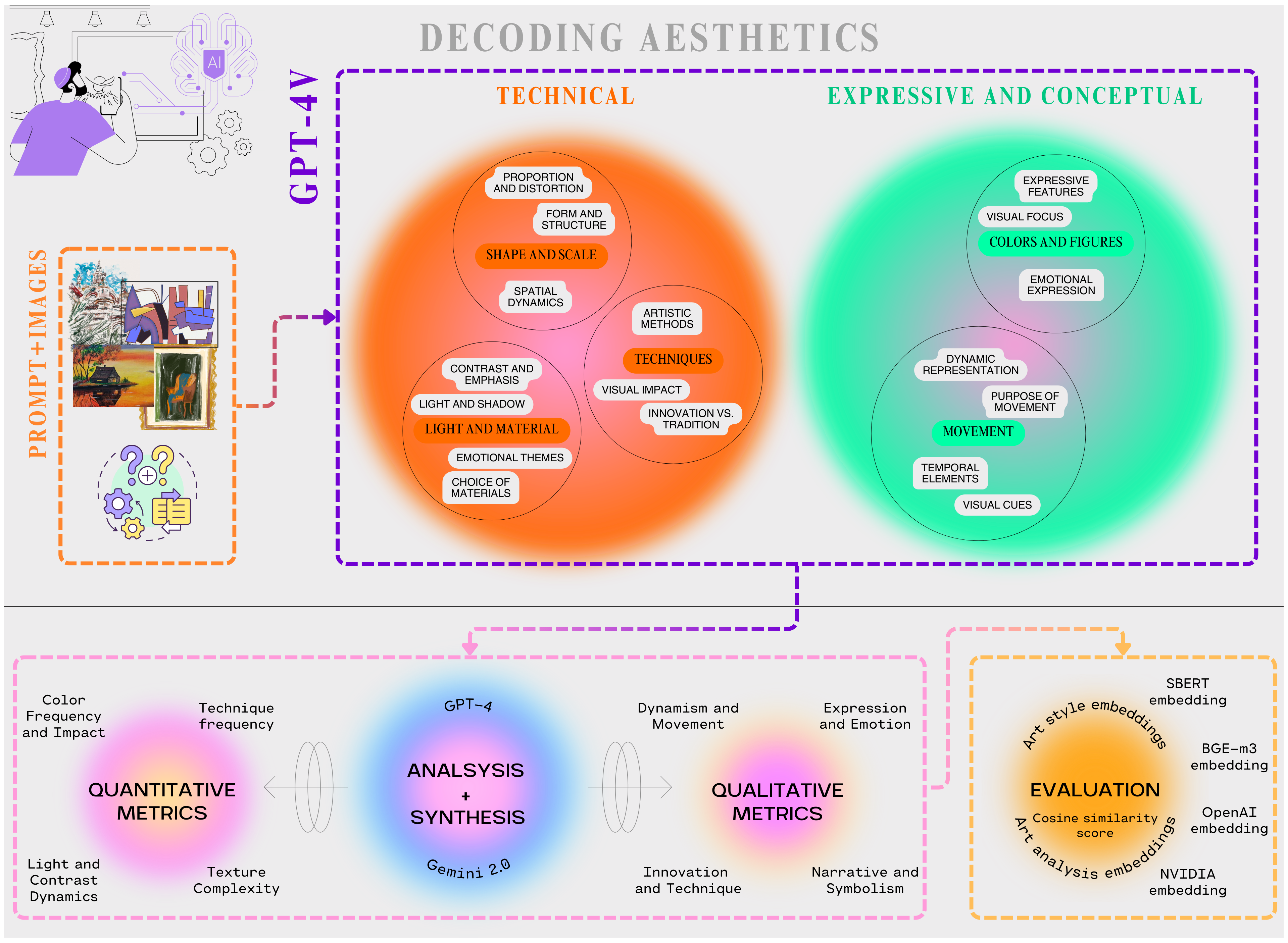}
    \caption{Illustration of our analysis framework for decoding aesthetics, which integrates both technical and expressive features of digitized artworks. The analysis process begins by submitting the artwork image along with eight predefined technical and conceptual questions to the GPT-4V API. The responses are subsequently processed by GPT-4 and Gemini 2.0 to extract and synthesize insights using both qualitative and quantitative art metrics. To evaluate the results, we compute the cosine similarity score between the text embeddings of art styles and the synthesized analysis results, leveraging four embedding models: SBERT, BGE-m3, OpenAI, and NVIDIA's NV-Embed-v2.}
    \label{fig:fig2}
\end{figure*}
\subsection{Proposed art analysis criteria}
\begin{figure*}[!ht] 
    \centering
    \includegraphics[width=\textwidth]{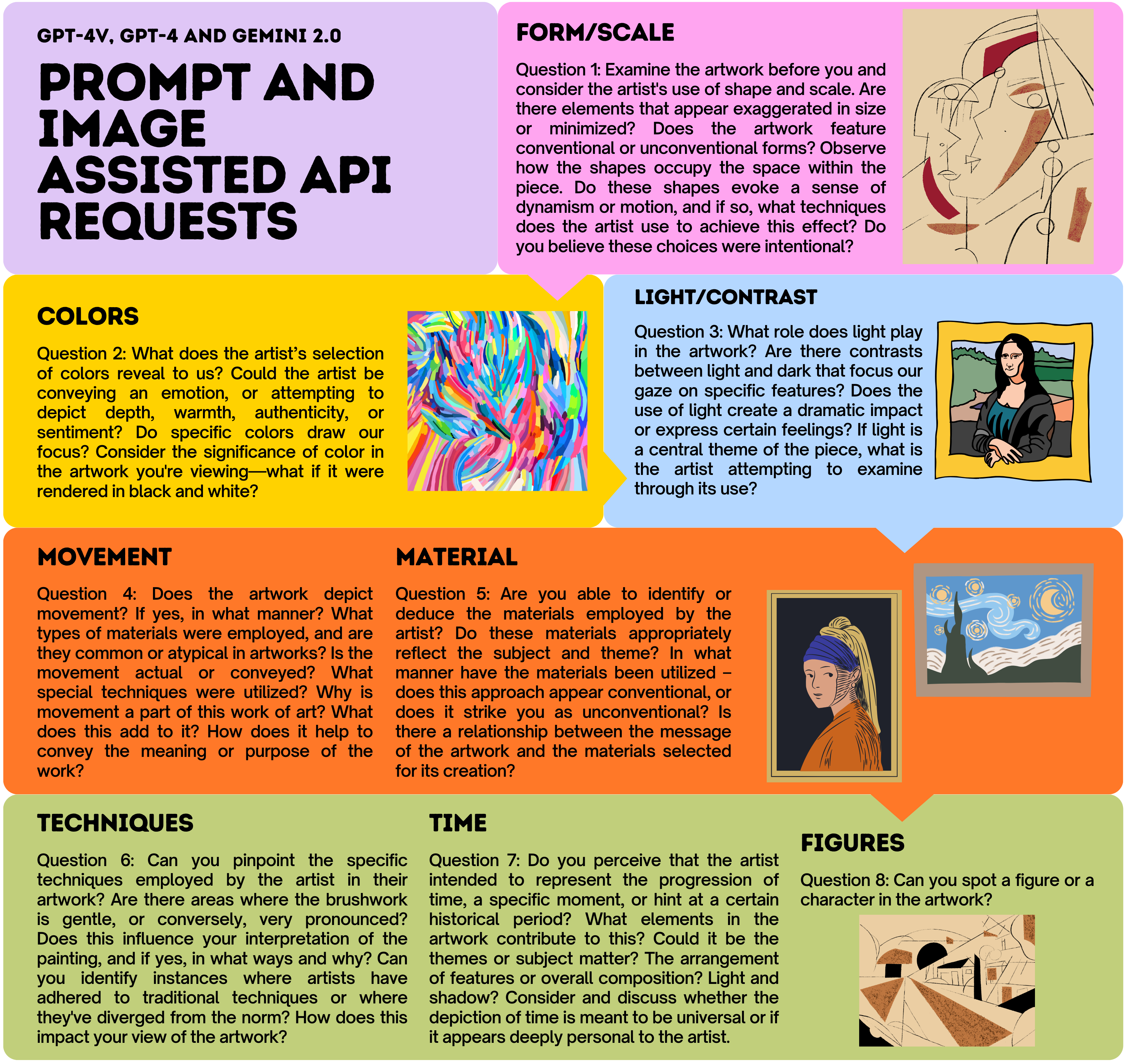}
    \caption{Demonstration of the technical and expressive range of questions we employed in our methodology to analyze the artworks, as incorporated into the API request prompts. Questions 1-7~\protect\cite{hodge2024elements} had been sourced carefully to guarantee that our analysis framework comply with the art assessment guidelines and expertise.}
    \label{fig:fig3}
\end{figure*}
Figure \ref{fig:fig2} illustrates the concept and structure used in our analysis. We evaluated the artworks from two distinct yet related perspectives: technical and conceptual. Each perspective encompasses a set of corresponding criteria, as detailed in Figure \ref{fig:fig2}. Technical analysis involves a broad range of criteria, including shape and scale, light, material, and the techniques employed by the artists. Expressive and conceptual analysis includes assessing the artworks based on their use of color, figures, objects, and movement.

To facilitate this analysis, we utilized GPT-4V. We sent API requests to the model, uploading a digitized image of each artwork along with a prompt consisting of eight questions~\cite{hodge2024elements} (Figure \ref{fig:fig3}). These questions covered the assessment groups mentioned above. More granularly, our analysis considered proportion and distortion, form and structure, spatial dynamics, artistic methods, visual impact, contrast, choice of material, emotional themes, light and shadow, visual focus and cues, and other critical criteria, as shown in Figure \ref{fig:fig2}. The responses from GPT-4V were then cleaned and processed before undergoing analysis and synthesis using GPT-4 and Gemini 2.0. This process allowed us to quantify both qualitative and quantitative metrics crucial for art evaluation, as presented in Figure \ref{fig:fig2}.

\section{Results}
Going forward, we will incorporate figures accessible online as supplementary material, designated with the prefix "S". The following is a structured presentation of the key findings of our study.

\subsection{Form and scale}
Figure \ref{fig:fig4S} depicts the distribution of the form types across different years. Natural forms have consistently been predominant throughout the observed period, maintaining a higher representation compared to geometric forms. In recent years, however, geometric forms have become increasingly prominent. Additionally, irregular forms have gained popularity more recently, surpassing the contributions of regular forms, which were previously more dominant across various periods. Figure \ref{fig:fig5S} illustrates the distribution of scale types among artists. It shows a preference for large or oversized scales in their work. Notably, there has been a recent increase in popularity for smaller or reduced scales among different artworks.

\subsection{Color}
We analyzed the color distribution in artworks to discern trends in emotional themes or color tones reflected by their palettes. Figure \ref{fig:fig6S} displays the temporal distribution of colors in our dataset. Certain colors have been predominantly used across various periods, while shades such as copper have only recently begun to emerge in our samples. Furthermore, Figure \ref{fig:fig7S} reveals that monochromatic tones were more prevalent in earlier periods but have seen a decline in popularity more recently with muted tones becoming more dominant. Additionally, we identified a correlation between color palettes and emotional themes, further interpreted through other elements like shapes and forms. Figure \ref{fig:fig8S} displays the emotional themes identified in our analysis. We successfully identified over 20 distinct emotional themes, with positive and neutral sentiments prevailing, as illustrated in Figure \ref{fig:fig9S}.

\subsection{Light, shadow and contrast}
Figure \ref{fig:fig10S} represents the distribution of light and contrast compositional elements. We aimed to explore the relationship between light, contrast, and their intended purposes in artworks. As demonstrated, most artists employ light and contrast to enhance highlights, depth, and textures. Similarly, the representation of emotions increased substantially during the 19th and 20th centuries. Conversely, elements such as faces and details, while present throughout the entire period, remained comparatively stable in their representation. 

The distribution of lighting and contrast effects is presented in Figure \ref{fig:fig11S}. The most prominent trend observed are the use of contrast together with shadow. Chiaroscuro was represented relatively consistently from the 17th century onwards. However, high contrast appeared to gain more popularity as the lighting and contrast effect more recently. In addition to the above, we also investigated the distribution of light types employed in artworks as illustrated in Figure \ref{fig:fig12S}. Diffused and soft lights were the most striking trend observed. Dappled light gained showed to prevail between 1860s to 1930s, while direct becoming prevalent more recently. Figure \ref{fig:fig13S} represents the distribution of the purposes of light, contrast, and shadow across top six art styles. As shown, artists have mainly used these elements to emphasize and draw attention. However, separating elements had been more prevalent in the surrealism compared to other other styles as one of the applications of these elements.

\subsection{Movement and material}
Figure \ref{fig:fig14S} illustrates the distribution of representations of movement. The results suggest that most of the artists have conveyed the movement rather than implying or suggesting it actually. Additionally, the findings indicate that most recent artworks have eschewed overt literal representations of movement in their compositions. The distribution of mediums and materials is presented in Figure \ref{fig:fig15S}. As shown, oil together canvas are the most dominant material types. Paper and ink were also frequently used, particularly in the 16th and 17th centuries, although to a lesser extent than oil and canvas. Wood was a prevalent material in the 16th and 17th centuries, with its use declining in later periods. Notably, acrylic emerged in the 20th century, reflecting the introduction of new synthetic materials in art production.

\subsection{Techniques}
Figure  \ref{fig:fig16S} represents the distribution of techniques and textures. The data reveal significant temporal variations. We found that blending and layering had been dominant techniques employed by artists. Smooth brushstrokes saw a significant increase in the late 19th and early 20th centuries, coinciding with the rise of Impressionism and related movements. Fine lines also exhibited a notable increase during the 19th and 20th centuries, though less pronounced than blending. Crosshatching showed a fluctuating trend from the 17th century onwards exhibiting a decline in the 20th century, while scarping and pointillism began to emerge.

\subsection{Time and figures}
Figure \ref{fig:fig17S} depicts the distribution of times of day and seasons in artworks. The data show that morning and afternoon settings have been more commonly portrayed compared to other times of the day, with this trend remaining relatively consistent across various periods. Recently, however, there has been an increase in artists opting for nighttime in their works. The distribution of the type of figures is illustrated in Figure \ref{fig:fig18S}. 
Human Figures consistently dominated representations throughout the entire period. Abstract and mythological figures appeared less frequently, with minor fluctuations in their portrayal. Meanwhile, fictional characters have gained increasing attention in more recent years.

\section{Evaluation}
In order to evaluate the accuracy of our analytical methodology, we compared our results with established descriptions of the ground-truth art styles. We transitioned from traditional topic modeling approaches, such as Latent Dirichlet Allocation (LDA)~\cite{blei2003latent}, to more advanced text embedding techniques to enhance the semantic accuracy of our analyses. Specifically, we utilized four state-of-the-art embedding models to transform textual data into meaningful, comparable representations. These models included Sentence-BERT (SBERT)~\cite{reimers-2019-sentence-bert}, BGE-M3~\cite{multim3}, OpenAI's text-embedding-3-small\footnote{\url{https://platform.openai.com/docs/guides/embeddings/}} and NVIDIA's NV-Embed-v2~\cite{lee2024nv}.

The results of the artwork analysis and the corresponding descriptions of known art styles were encoded into embeddings using the four models mentioned above. Specifically, for the SBERT embeddings, we utilized the "all-mpnet-base-v2" model \cite{thakur-2020-AugSBERT}, which is fine-tuned to capture sentence-level semantic meanings effectively. All evaluations were conducted on an Apple MacBook M3 Pro with a 14-core GPU, except for the final model. For the evaluation with NV-Embed-v2, we utilized four RTX A6000 GPUs, each equipped with 48 GB of GPU memory.

The core of our evaluation process involves computing the cosine similarity between the embeddings generated from the art analysis results and those derived from style-specific descriptions. This approach enables us to quantitatively assess the alignment between our analytical outputs and established art style descriptors. 
Based on the statistics presented in Table 1 (also shown in Figures \ref{fig:fig19S}-\ref{fig:fig23S}), our evaluation demonstrates significant variation in the cosine similarity scores across different embedding models and focus areas. Among the models, NVIDIA’s NV-Embed-v2 consistently exhibits the highest median scores across all focus areas, with particularly strong performance in Form/Scale (0.62), Movement (0.63), and Techniques (0.70). This suggests that NV-Embed-v2 embedding space captures stylistic features more effectively than other models, particularly in technical and dynamic aspects of artwork. BGE-m3 follows closely, achieving comparable median scores in most categories, especially in Light/Contrast (0.61), Movement (0.61), and Material (0.62). On the other hand, SBERT and OpenAI models generally yield lower similarity scores, indicating that their embeddings may be less specialized in distinguishing artistic styles based on our predefined criteria.
\begin{table}[!ht]
    \centering
    \begin{tabular}{lllllll}
        \hline
        \textbf{Model} & \textbf{Focus} & \textbf{Min} & \textbf{Q1} & \textbf{Med} & \textbf{Q3} & \textbf{Max} \\
        \hline
        \multirow{6}{*}{\textbf{S}BERT} & F/S & 0.18 & 0.39 & 0.44 & 0.48 & 0.81 \\
                               & CLR & 0.18 & 0.37 & 0.41 & 0.45 & 0.71 \\
                               & L/C & 0.17 & 0.37 & 0.41 & 0.45 & 0.73 \\
                               & MVT & 0.04 & 0.38 & 0.44 & 0.50 & 0.85 \\
                               & MAT & 0.10 & 0.37 & 0.44 & 0.51 & 0.78 \\
                               & TCH & 0.20 & 0.42 & 0.50 & 0.57 & 0.84 \\
        \hline
        \multirow{6}{*}{\textbf{B}GE-m3} & F/S & 0.45 & 0.59 & \textbf{0.62} & 0.64 & 0.79 \\
                            & CLR & 0.46 & 0.58 & \textbf{0.61} & 0.64 & 0.80 \\
                            & L/C & 0.41 & 0.59 & \textbf{0.61} & 0.64 & 0.78 \\
                            & MVT & 0.42 & 0.58 & 0.61 & 0.66 & 0.81 \\
                            & MAT & 0.45 & 0.59 & 0.62 & 0.65 & 0.81 \\
                            & TCH & 0.49 & 0.61 & 0.65 & 0.68 & 0.85 \\
        \hline
        \multirow{6}{*}{\textbf{O}penAI} & F/S & 0.14 & 0.34 & 0.39 & 0.43 & 0.72 \\
                                & CLR & 0.17 & 0.35 & 0.40 & 0.44 & 0.68 \\
                                & L/C & 0.16 & 0.33 & 0.37 & 0.41 & 0.66 \\
                                & MVT & 0.17 & 0.34 & 0.39 & 0.46 & 0.71 \\
                                & MAT & 0.15 & 0.37 & 0.43 & 0.48 & 0.82 \\
                                & TCH & 0.20 & 0.40 & 0.47 & 0.54 & 0.81 \\
        \hline
        \multirow{6}{*}{\textbf{N}VIDIA} & F/S & 0.30 & 0.54 & 0.61 & 0.69 & \textbf{0.98} \\
                                & CLR & 0.19 & 0.52 & 0.59 & 0.64 & \textbf{0.90} \\
                                & L/C & 0.18 & 0.45 & 0.52 & 0.56 & \textbf{0.84} \\
                                & MVT & 0.18 & 0.55 & \textbf{0.63} & 0.71 & \textbf{0.96} \\
                                & MAT & 0.41 & 0.58 & \textbf{0.64} & 0.71 & \textbf{0.90} \\
                                & TCH & 0.34 & 0.65 & \textbf{0.70} & 0.75 & \textbf{0.90} \\
        \hline
        \hline
        \multirow{6}{*}{\textbf{SBON*}} & TCH & 0.20 & 0.48 & \underline{\textbf{0.61}} & 0.68 & 0.90 \\
                                & MVT & 0.04 & 0.48 & 0.57 & 0.65 & 0.96 \\
                                & MAT & 0.10 & 0.44 & 0.56 & 0.63 & 0.90 \\
                                & F/S & 0.14 & 0.41 & 0.51 & 0.62 & 0.98 \\
                                & CLR & 0.17 & 0.40 & 0.48 & 0.60 & 0.90 \\
                                & L/C & 0.16 & 0.38 & 0.46 & 0.58 & 0.84 \\
        \hline
        \hline
    \end{tabular}
    \caption{Statistics for cosine similarity score between the embeddings of art styles and analysis results, across different embedding models and focus areas: F/S (Form/Scale), CLR (Colors), L/C (Light/Contrast), MVT (Movement), MAT (Material), and TCH (Techniques). Embedding models included: SBERT's all-mpnet-base-v2, Flag Embedding's BGE-m3, OpenAI's text-embedding-3-small and NVIDIA's NV-Embed-v2. \textbf{SBON} represents the aggregated summary statistics for the cosine similarity scores across all models, with focus areas sorted in descending order based on their median scores.}
    \label{tab:cosine_similarity_scores}
\end{table}
Aggregating across all models, the \textbf{SBON} (summary statistics) row highlights that Techniques (0.61 median) and Movement (0.57 median) emerge as the most consistently aligned features between analysis results and art style embeddings. This indicates that aspects related to brushwork, composition techniques, and movement patterns are more effectively captured by language models compared to material or light-based attributes. The relatively lower similarity scores in Material (0.40 median) suggest that existing embedding models may struggle with material representation, possibly due to the abstract nature of material descriptions in text form. The maximum scores across models also reveal strong performance in Colors (0.90), Form/Scale (0.98), and Movement (0.96), implying that some embeddings can achieve high stylistic alignment under optimal conditions. These findings highlight the strengths and weaknesses of different models, suggesting that a multi-model ensemble approach could enhance the robustness of aesthetic decoding in digital art analysis.
\section{Discussion}
Our study demonstrates the significant potential of \glspl{llm}, specifically GPT-4V, Gemini 2.0, and GPT-4, in revolutionizing the field of art analysis. By leveraging these advanced AI models, we were able to process and interpret over 15,000 artworks across 34 different art styles from 23 prominent artists spanning multiple historical periods. Our findings highlight the ability of \glspl{llm} to extract meaningful insights into the technical and expressive elements of art, providing a more objective and efficient approach to decoding artistic styles, visual elements, composition, and techniques.

Our analysis revealed a consistent dominance of natural forms throughout the observed periods, with a recent surge in geometric and irregular forms. The temporal distribution of colors demonstrated a dynamic interplay between tradition and innovation. The analysis of light, shadow, and contrast revealed how artists have used these elements to enhance depth, texture, and emotional expression in their works. The distribution of movement representations suggests a shift away from literal depictions of motion towards more suggestive and abstract approaches in recent artworks. The temporal variations in techniques and textures provide insights into the evolution of artistic styles and practices.


\textbf{Strengths}: Our study demonstrates the strengths of \glspl{llm} in processing and interpreting vast amounts of art-related data, generating insightful interpretations, and automating the analysis process. The ability of GPT-4V to analyze both visual and textual information allows for a comprehensive understanding of artworks, bridging the gap between computer vision and human interpretation. The integration of Gemini 2.0 and GPT-4 further enhances the analysis by enabling sophisticated synthesis and interpretation of the results.

\textbf{Limitations}: However, our study also has limitations. The subjective nature of art interpretation means that even the most advanced \gls{ai} models may not fully capture the nuances of human perception and emotional response to art. The accuracy and depth of the analysis are also dependent on the quality and diversity of the available data. While our dataset is extensive, it may not fully represent the entire spectrum of artistic styles and periods. Additionally, the analysis relies on the predefined criteria and questions, which may not cover all aspects of art analysis and could introduce biases.

\section{Conclusion and Future Work}

Our findings demonstrate that our approach to decoding the authentic nuances of artworks can be effectively addressed using \glspl{llm}. This method not only reveals patterns and trends in the evolution of artistic styles, visual elements, composition, and techniques over time but also generates insightful interpretations. Additionally, it provides a more objective and efficient means of art analysis, complementing traditional methods and offering a valuable tool for art historians, researchers, and enthusiasts. 

Future research could explore further advancements in \gls{llm} techniques and their applications in a broader range of artworks. Investigating the potential of \glspl{llm} in other aspects of art analysis—such as identifying forgeries, understanding the influence of social and cultural factors on art, and predicting future trends could deepen our understanding of art and its evolution. Furthermore, the development of more interactive and user-friendly interfaces for LLM-based art analysis could make these powerful tools more accessible to students, educators, and art enthusiasts.
\clearpage
\section*{Acknowledgments}
This research was funded in whole by the Luxembourg National Research Fund (FNR), grant reference 16326754. For the purpose of open access, and in fulfillment of the obligations arising from the grant agreement, the author has applied a Creative Commons Attribution 4.0 International (CC BY 4.0) license to any Author Accepted Manuscript version arising from this submission.
\section*{Ethical Statement}
There are no ethical issues.

\bibliographystyle{named}
\bibliography{ijcai25}
\clearpage

\counterwithin{figure}{section}
\setcounter{figure}{0}
\renewcommand{\thefigure}{S\arabic{figure}}
\section*{Supplementary Figures}
This section provides supplementary figures that support the findings and discussions presented in the main body of our paper. Due to the interactive nature of the data visualizations, these figures are also available online at ~\href{https://cognartive.github.io/}{\textcolor{blue}{https://cognartive.github.io/}}.
\begin{figure*}[!ht] 
    \centering
    \includegraphics[width=\textwidth]{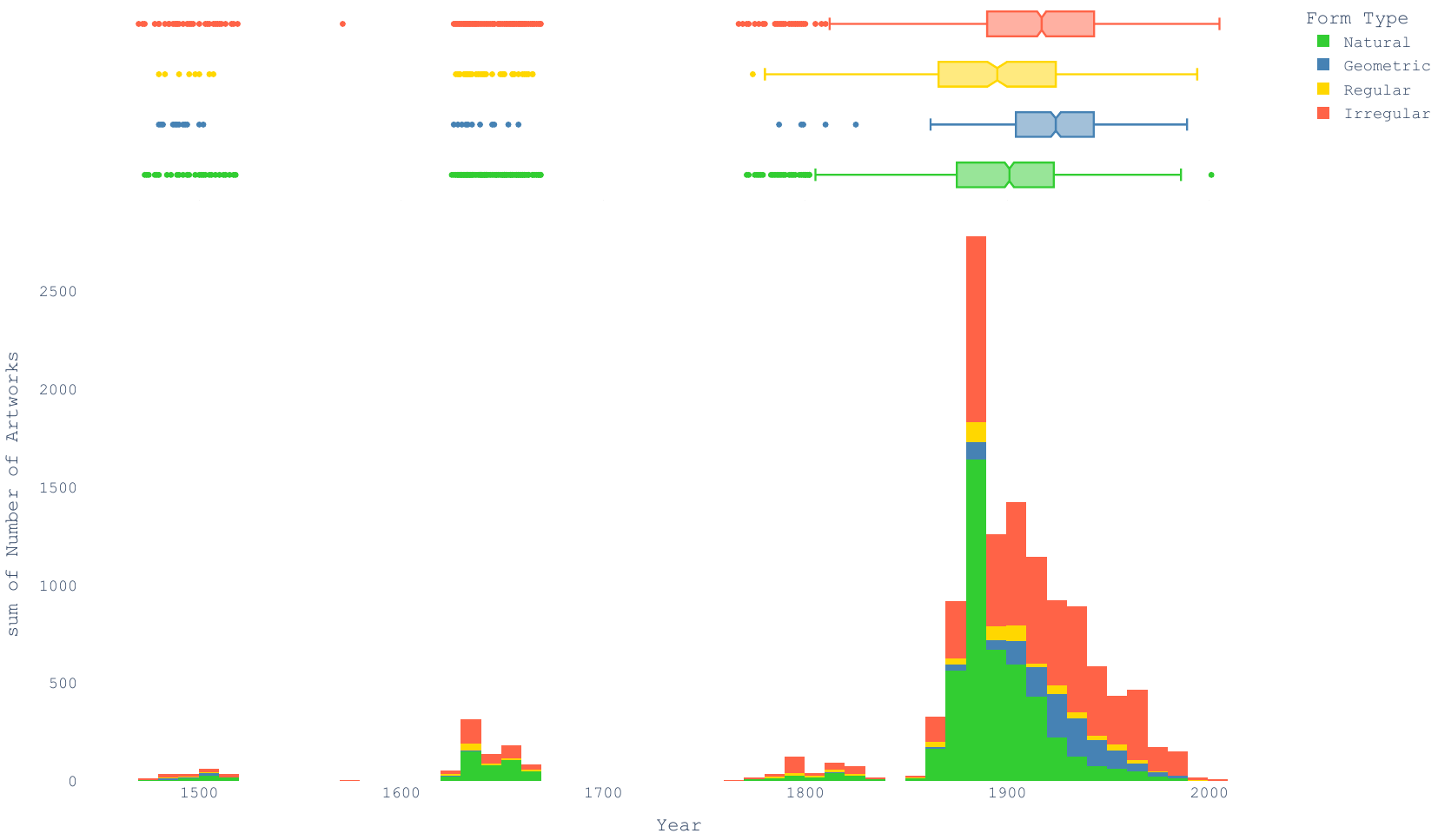}
    \caption{Distribution of form types in artworks (Natural, Geometric, Regular, and Irregular) across different years. The chart shows the cumulative number of artworks for each form type.
    }
    \label{fig:fig4S}
\end{figure*}

\begin{figure*}[!ht] 
    \centering
    \includegraphics[width=\textwidth]{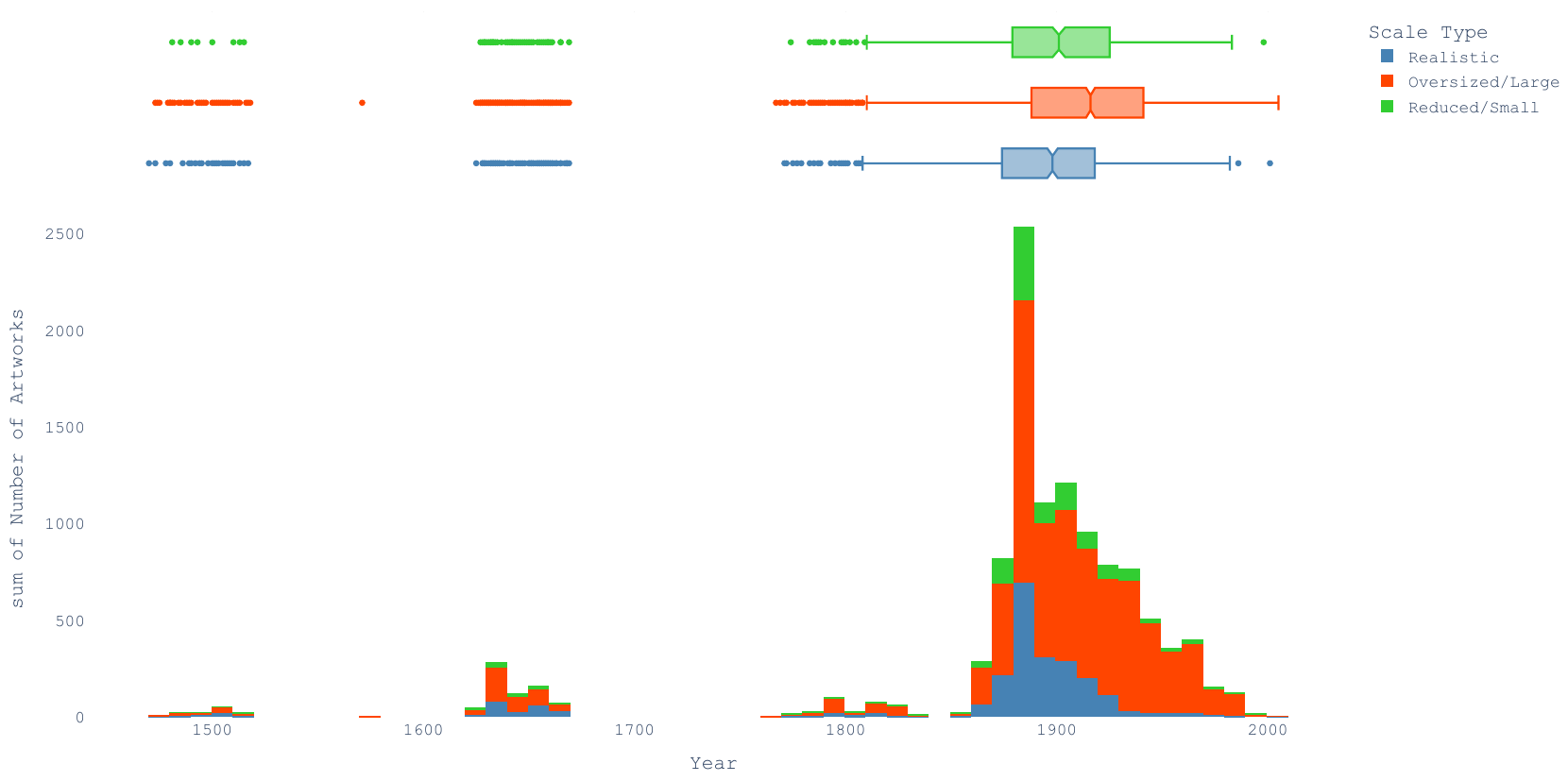}
    \caption{Distribution of scale types in artworks (Realistic, Oversized/Large, and Reduced/Small) across different years. The chart shows the cumulative number of artworks for each scale type.}
    \label{fig:fig5S}
\end{figure*}

\begin{figure*}[!ht] 
    \centering
    \includegraphics[angle=90,origin=c,width=\textwidth,height=0.8\textheight,keepaspectratio]{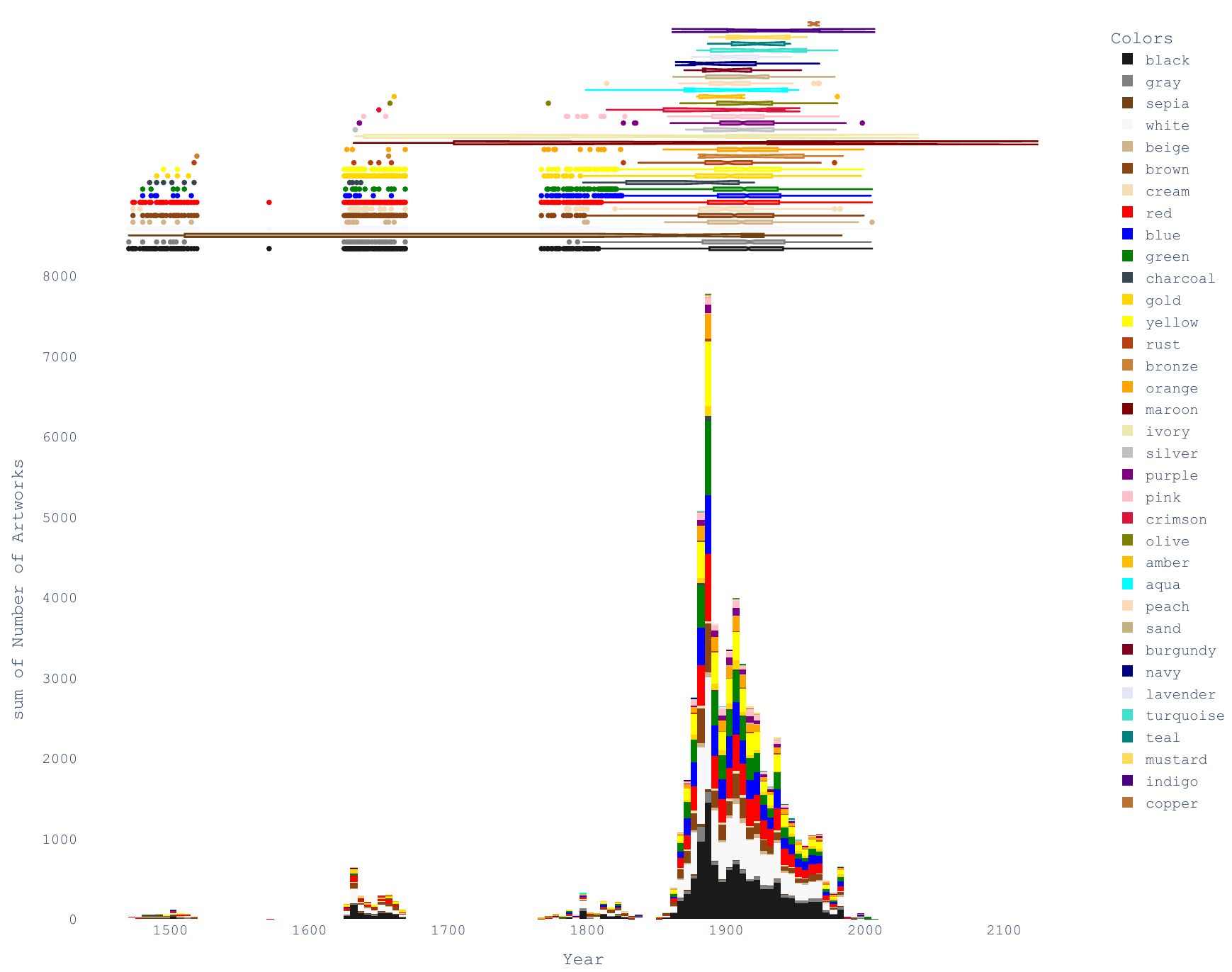}
    \caption{Distribution of color palettes employed in artworks across different years.}
    \label{fig:fig6S}
\end{figure*}

\begin{figure*}[!ht] 
    \centering
    \includegraphics[angle=90,origin=c,width=\textwidth,height=0.7\textheight,keepaspectratio]{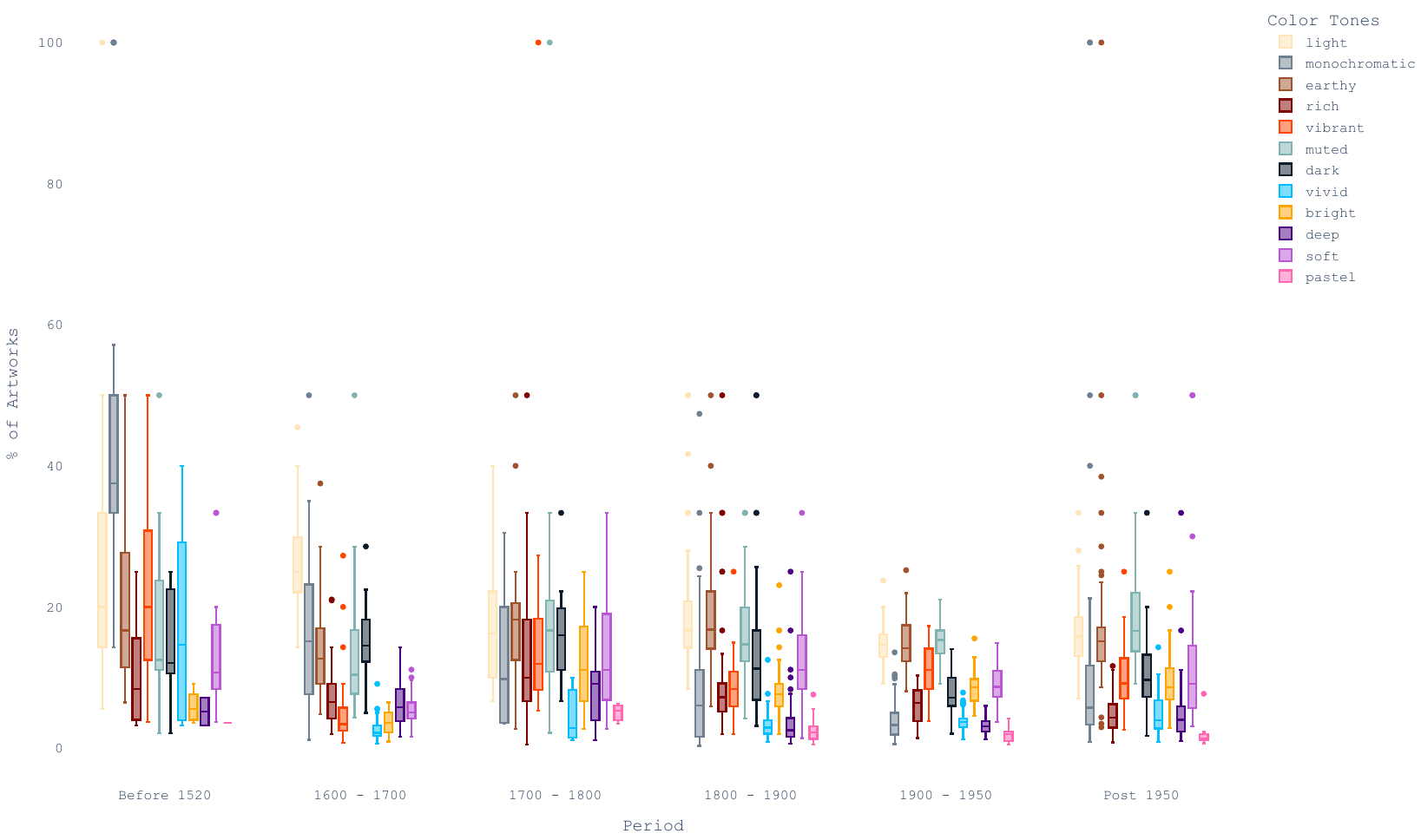}
    \caption{Distribution of color tones in artworks across different historical periods. The box plots represent the percentage of each color tone used in artworks.}
    \label{fig:fig7S}
\end{figure*}

\begin{figure*}[!ht] 
    \centering
    \includegraphics[width=\textwidth]{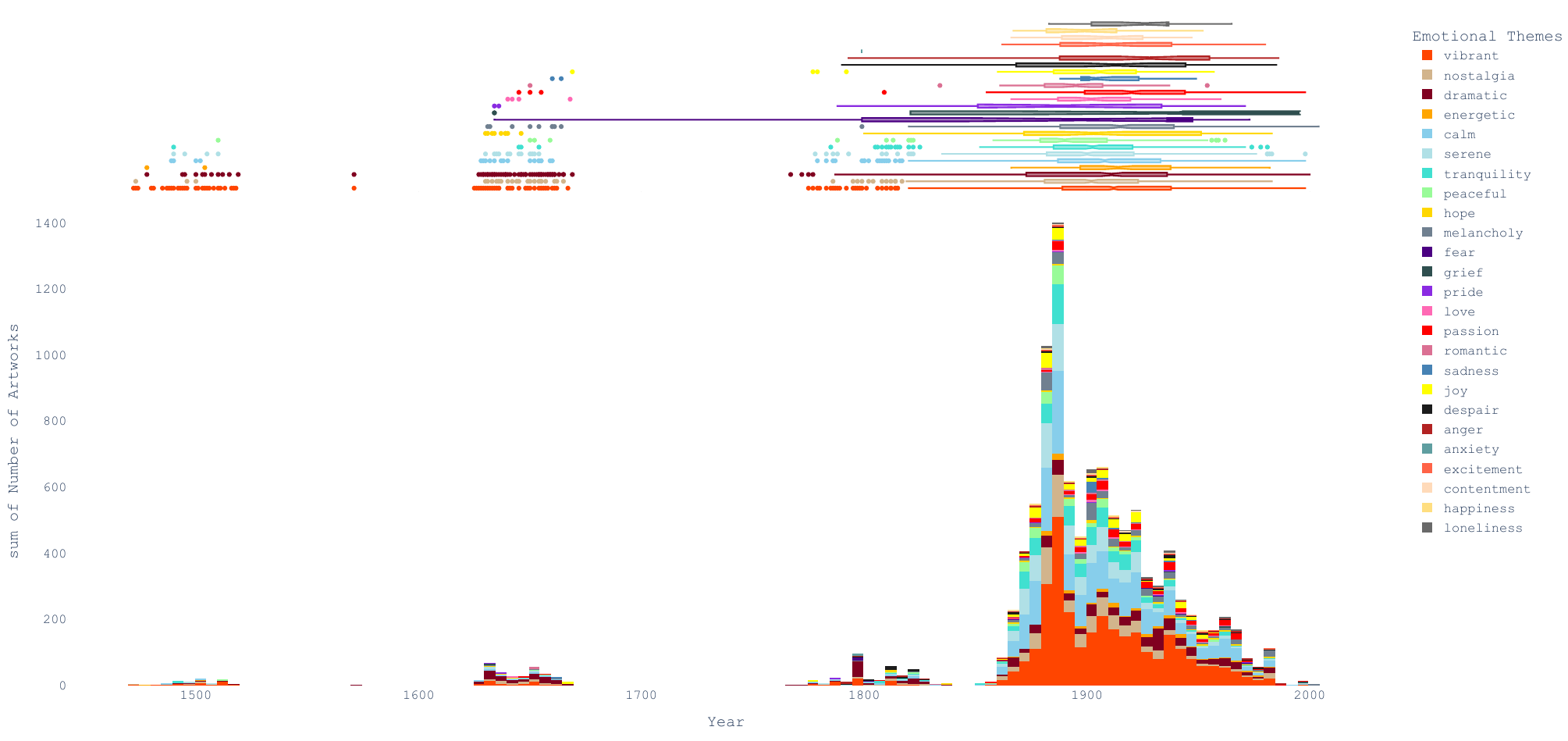}
    \caption{Distribution of emotional themes interpreted from artworks across different years.}
    \label{fig:fig8S}
\end{figure*}

\begin{figure*}[!ht] 
    \centering
    \includegraphics[width=\textwidth]{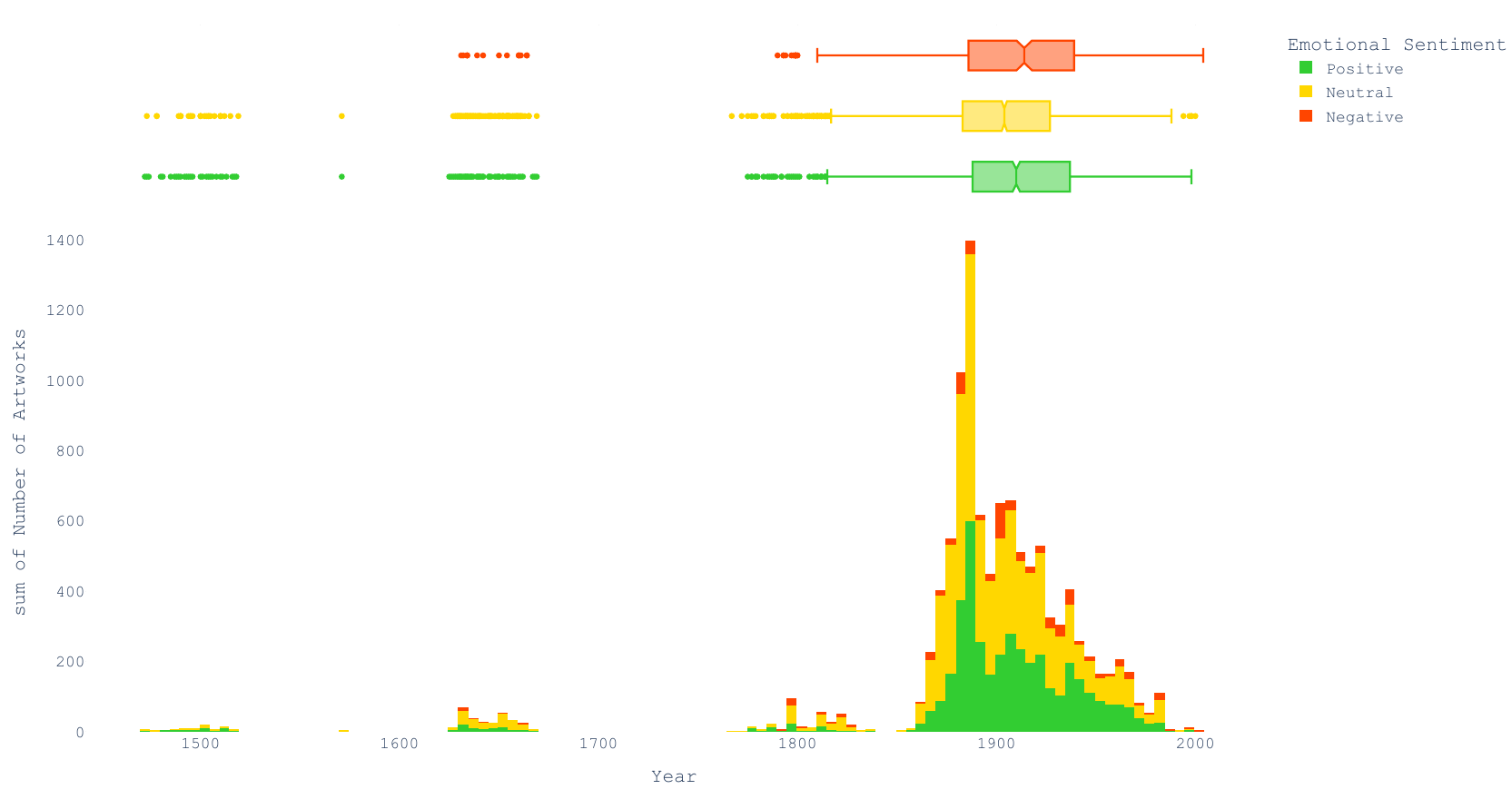}
    \caption{Distribution of emotional sentiments (Positive, Neutral and Negative) interpreted from artworks across different years.}
    \label{fig:fig9S}
\end{figure*}

\begin{figure*}[!ht] 
    \centering
    \includegraphics[width=\textwidth]{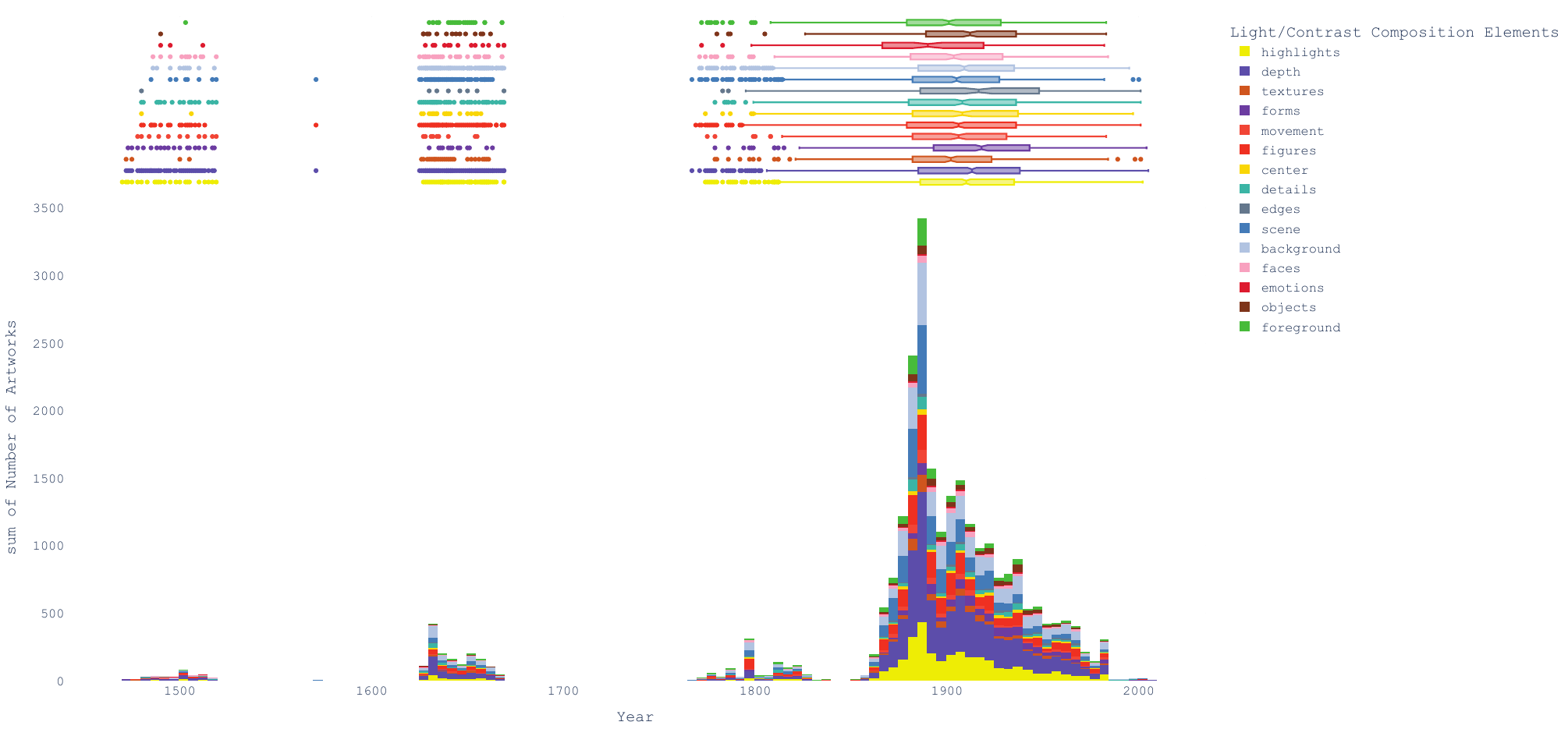}
    \caption{Distribution of light and contrast composition elements used in artworks across different years. This chart highlights the main target of these elements used by artists in their works.}
    \label{fig:fig10S}
\end{figure*}

\begin{figure*}[!ht] 
    \centering
    \includegraphics[width=\textwidth]{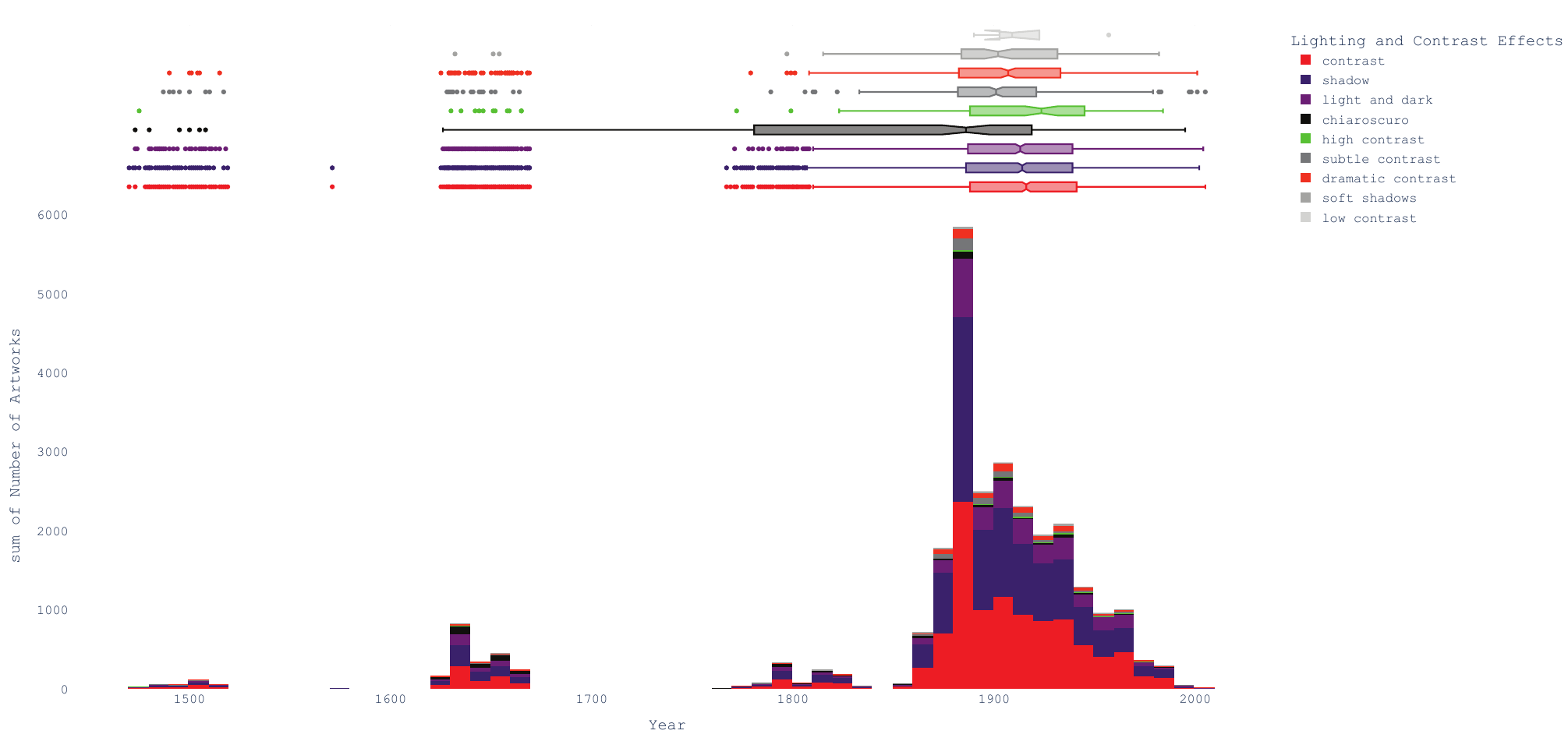}
    \caption{Distribution of lighting and contrast effects used in artworks across different years.}
    \label{fig:fig11S}
\end{figure*}

\begin{figure*}[!ht] 
    \centering
    \includegraphics[width=\textwidth]{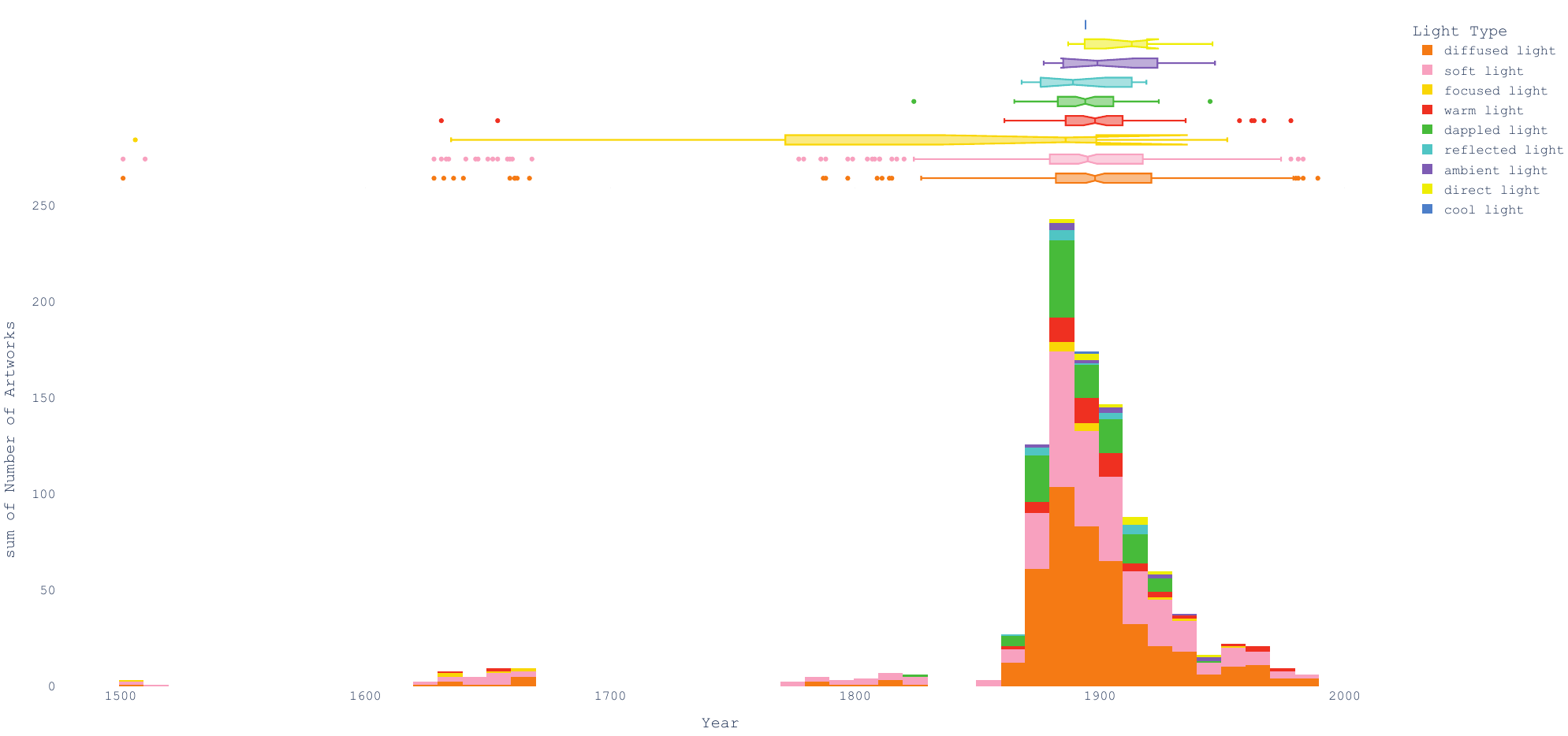}
    \caption{Distribution of the light types used in artworks across different years.}
    \label{fig:fig12S}
\end{figure*}

\begin{figure*}[!ht] 
    \centering
    \includegraphics[width=\textwidth]{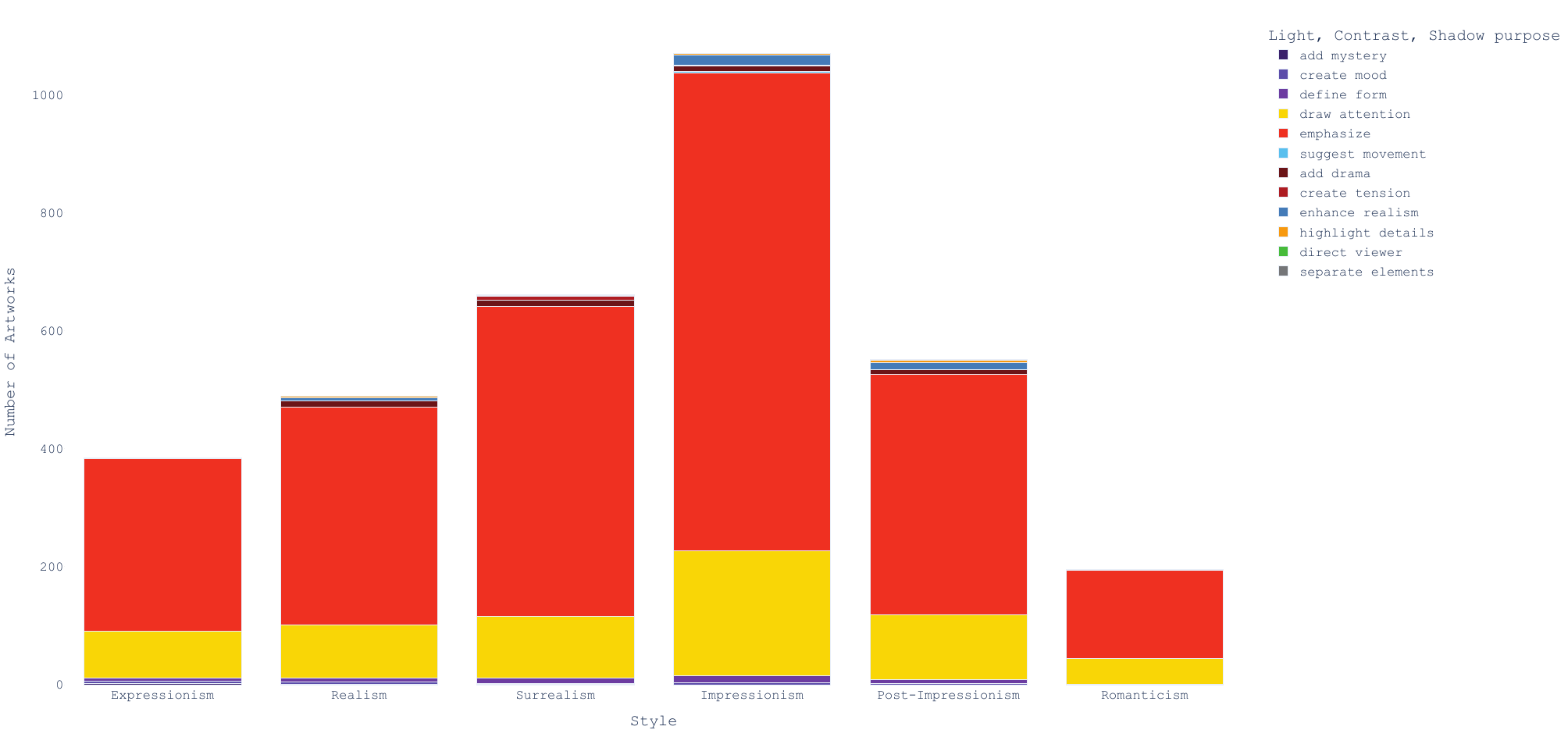}
    \caption{Distribution of the purposes of application of light, contrast and shadow in artworks across different art styles.}
    \label{fig:fig13S}
\end{figure*}

\begin{figure*}[!ht] 
    \centering
    \includegraphics[width=\textwidth]{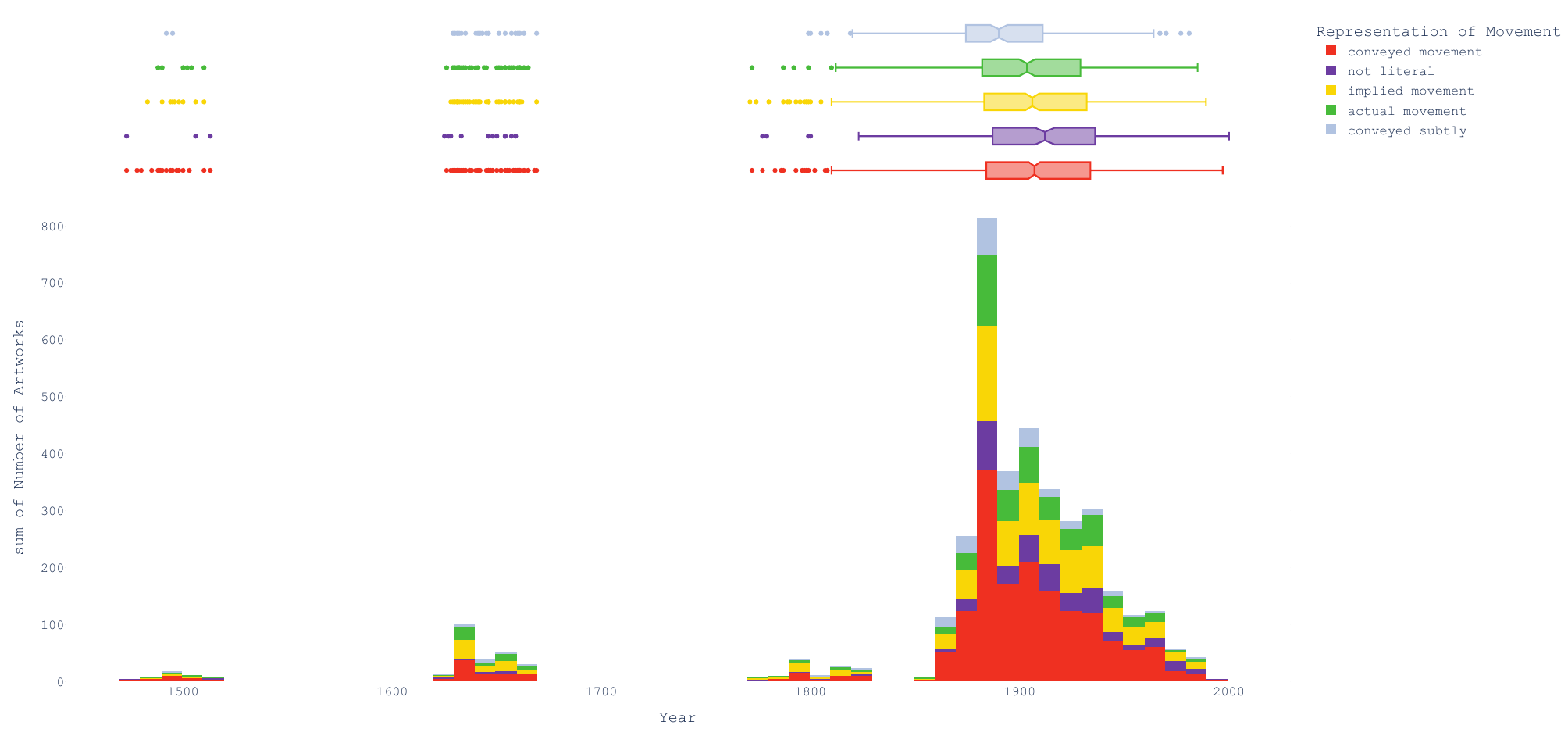}
    \caption{Distribution of the representations of movements interpreted from artworks across different years.}
    \label{fig:fig14S}
\end{figure*}

\begin{figure*}[!ht] 
    \centering
    \includegraphics[width=\textwidth]{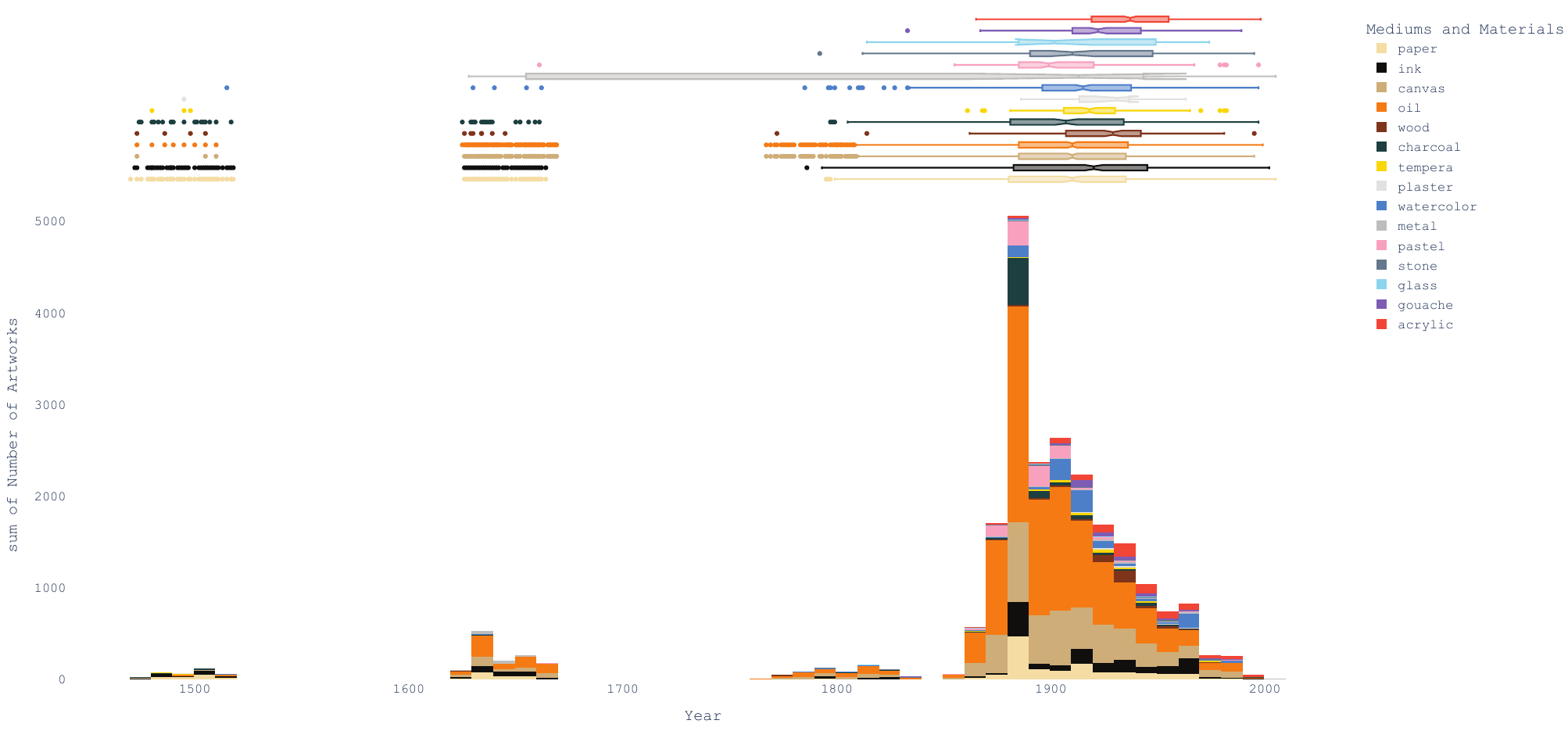}
    \caption{Distribution of the mediums and materials types employed in artworks across different years.}
    \label{fig:fig15S}
\end{figure*}

\begin{figure*}[!ht] 
    \centering
    \includegraphics[width=\textwidth]{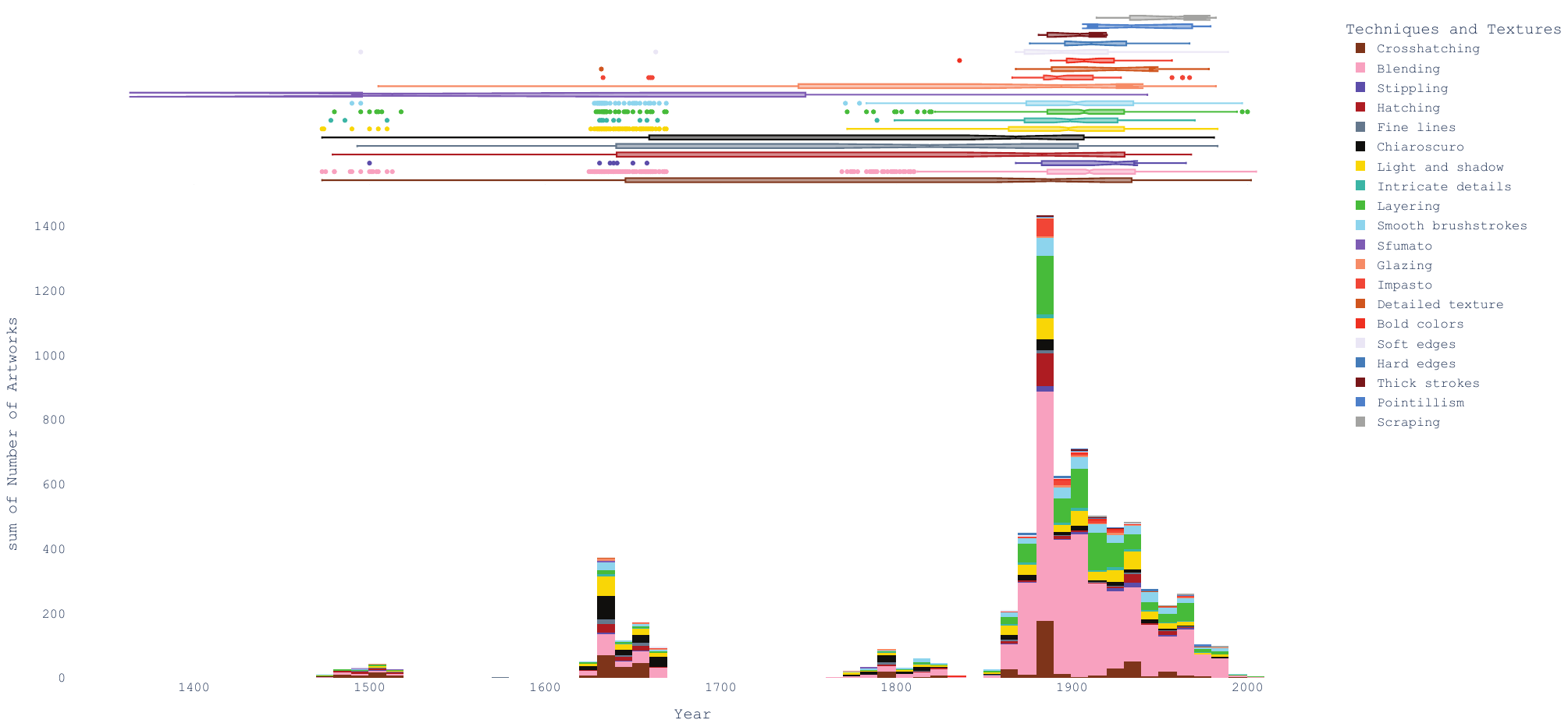}
    \caption{Distribution of the technique and texture types utilized in artworks across different years.}
    \label{fig:fig16S}
\end{figure*}

\begin{figure*}[!ht] 
    \centering
    \includegraphics[width=\textwidth]{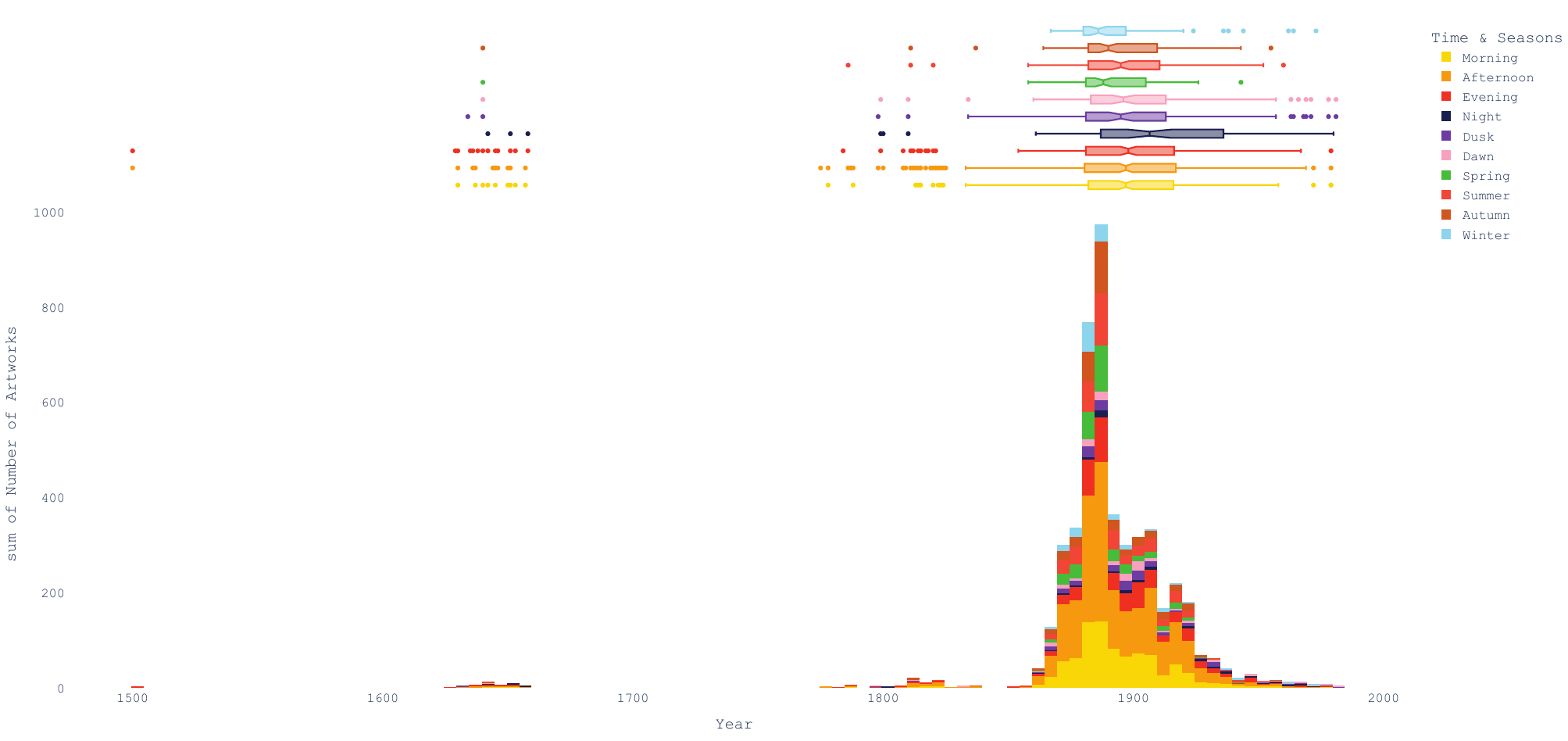}
    \caption{Distribution of the time of the day and season types interpreted from artworks across different years.}
    \label{fig:fig17S}
\end{figure*}

\begin{figure*}[!ht] 
    \centering
    \includegraphics[width=\textwidth]{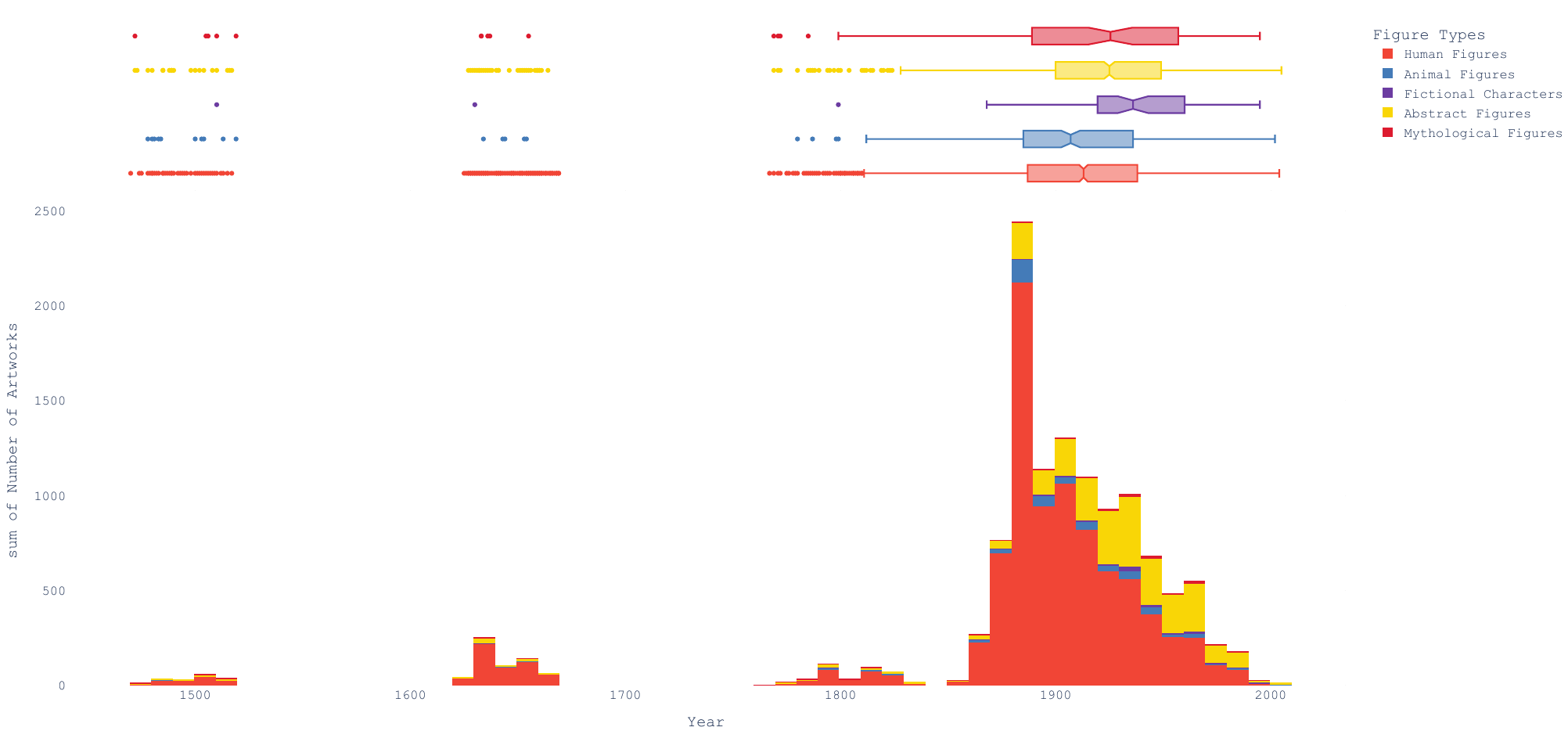}
    \caption{Distribution of the figure types interpreted from artworks across different years.}
    \label{fig:fig18S}
\end{figure*}

\begin{figure*}[!ht] 
    \centering
    \includegraphics[angle=90,origin=c,width=\textwidth,height=0.67\textheight,keepaspectratio]{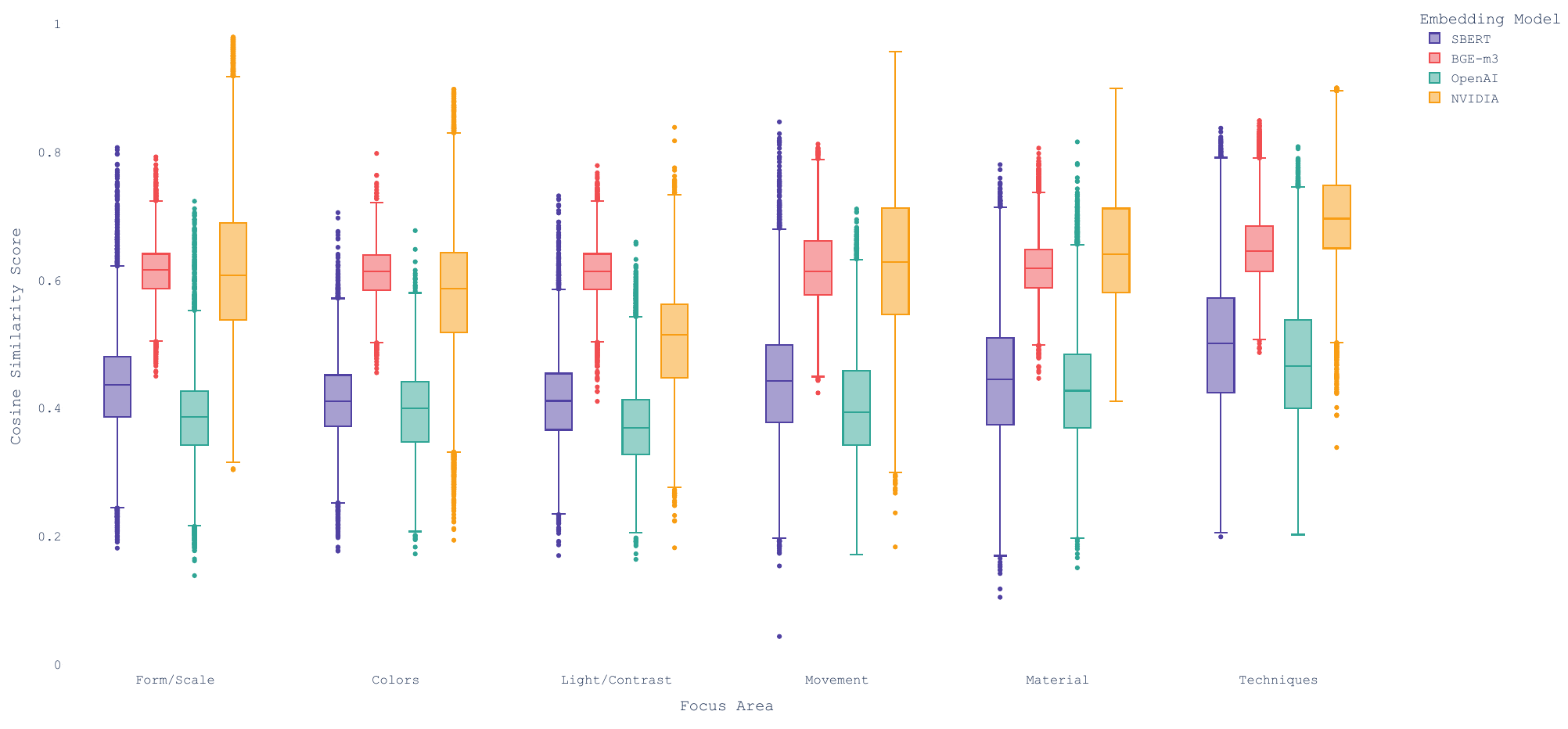}
    \caption{Distribution of cosine similarity scores for six distinct focus areas—Form/Scale, Colors, Light/Contrast, Movement, Material, and Techniques—across four embedding models: SBERT, BGE-m3, OpenAI, and NVIDIA. The NVIDIA model consistently demonstrates higher median similarity scores with wider score distributions, suggesting greater sensitivity to variations. In contrast, the SBERT and OpenAI models exhibit lower median scores and tighter distributions, indicating more conservative similarity assessments.}
    \label{fig:fig19S}
\end{figure*}

\begin{figure*}[!ht] 
    \centering
    \includegraphics[angle=90,origin=c,width=\textwidth,height=0.7\textheight,keepaspectratio]{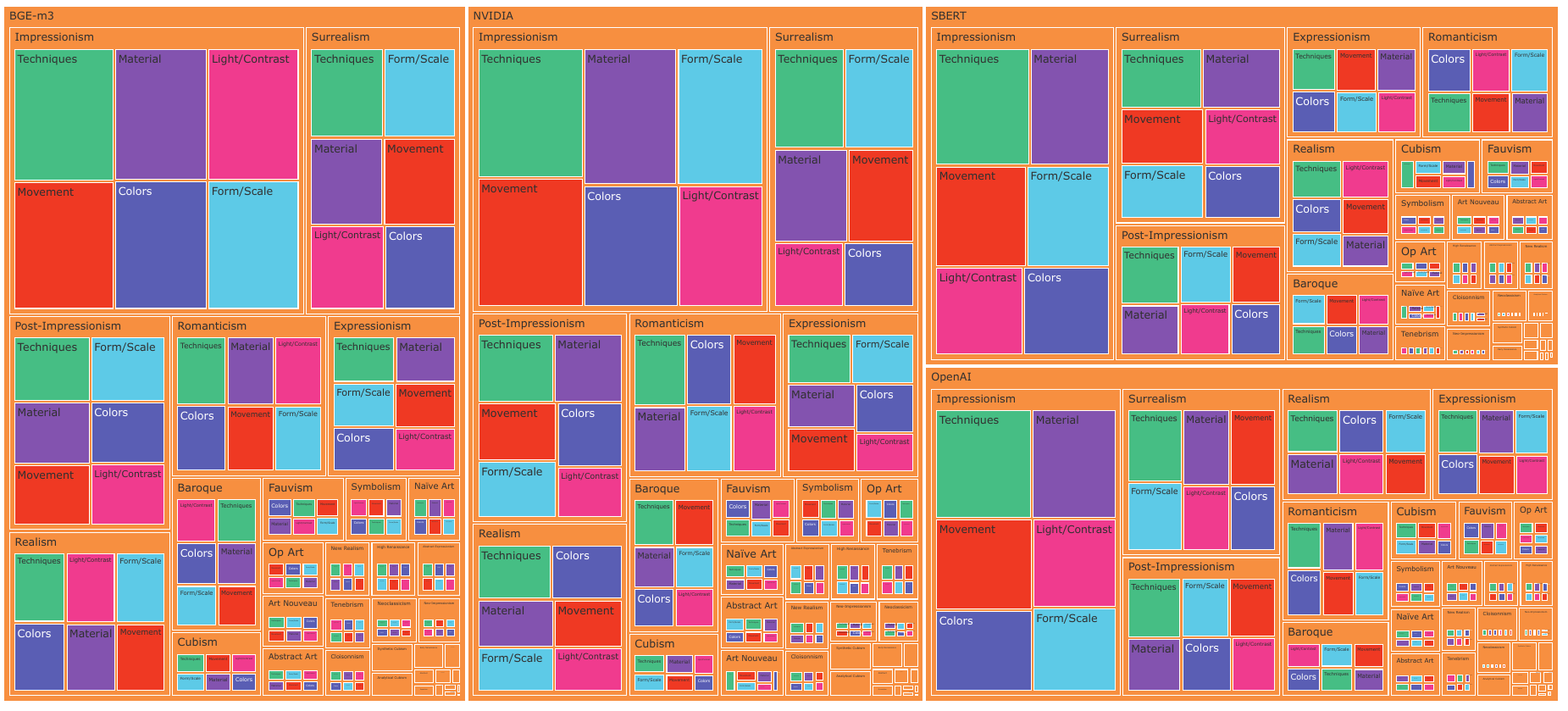}
    \caption{Treemap of the cumulative cosine similarity scores per embedding model, art style and analysis focus areas.}
    \label{fig:fig20S}
\end{figure*}

\begin{figure*}[!ht] 
    \centering
    \includegraphics[angle=90,origin=c,width=\textwidth,height=0.7\textheight,keepaspectratio]{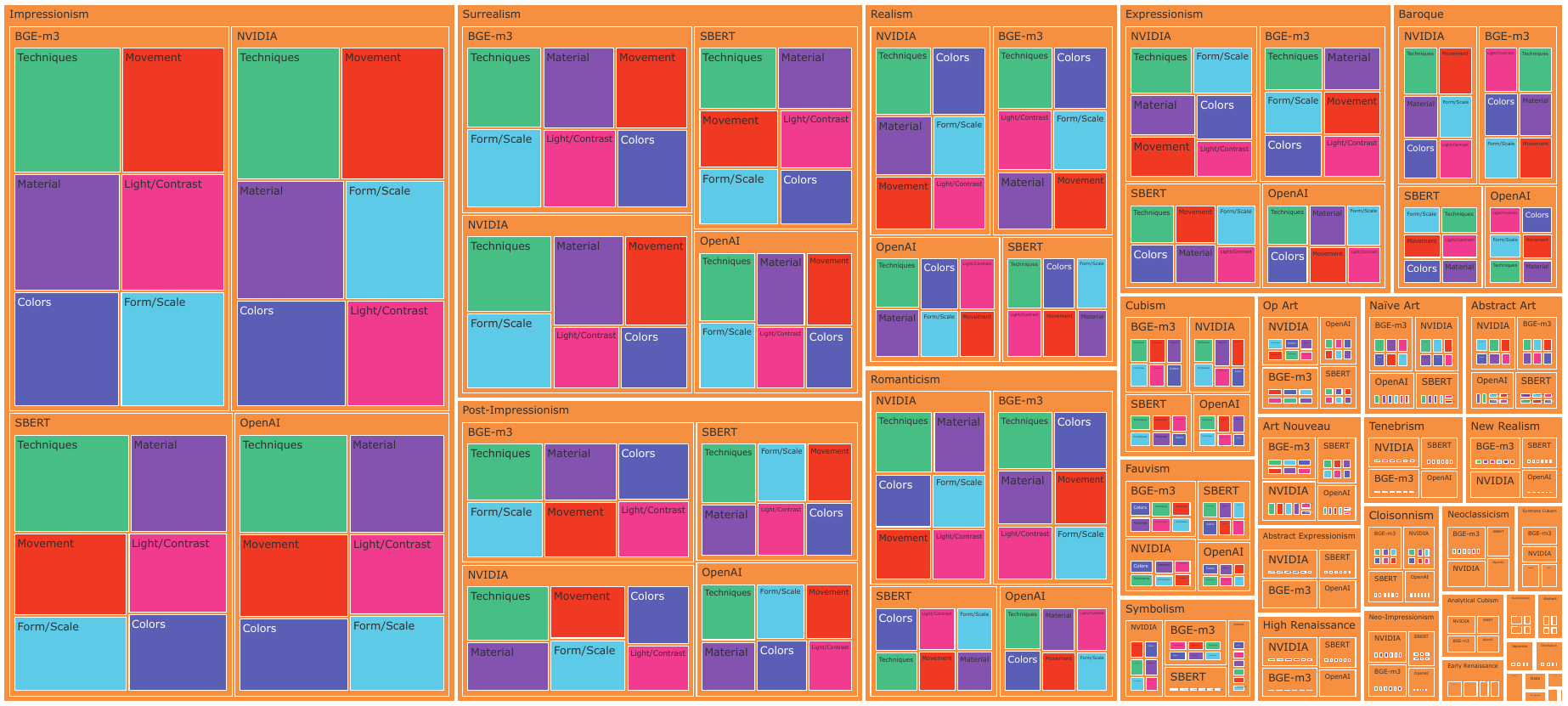}
    \caption{Treemap of the cumulative cosine similarity scores per art style, embedding model and analysis focus areas.}
    \label{fig:fig21S}
\end{figure*}

\begin{figure*}[!ht] 
    \centering
    \includegraphics[angle=90,origin=c,width=\textwidth,height=0.7\textheight,keepaspectratio]{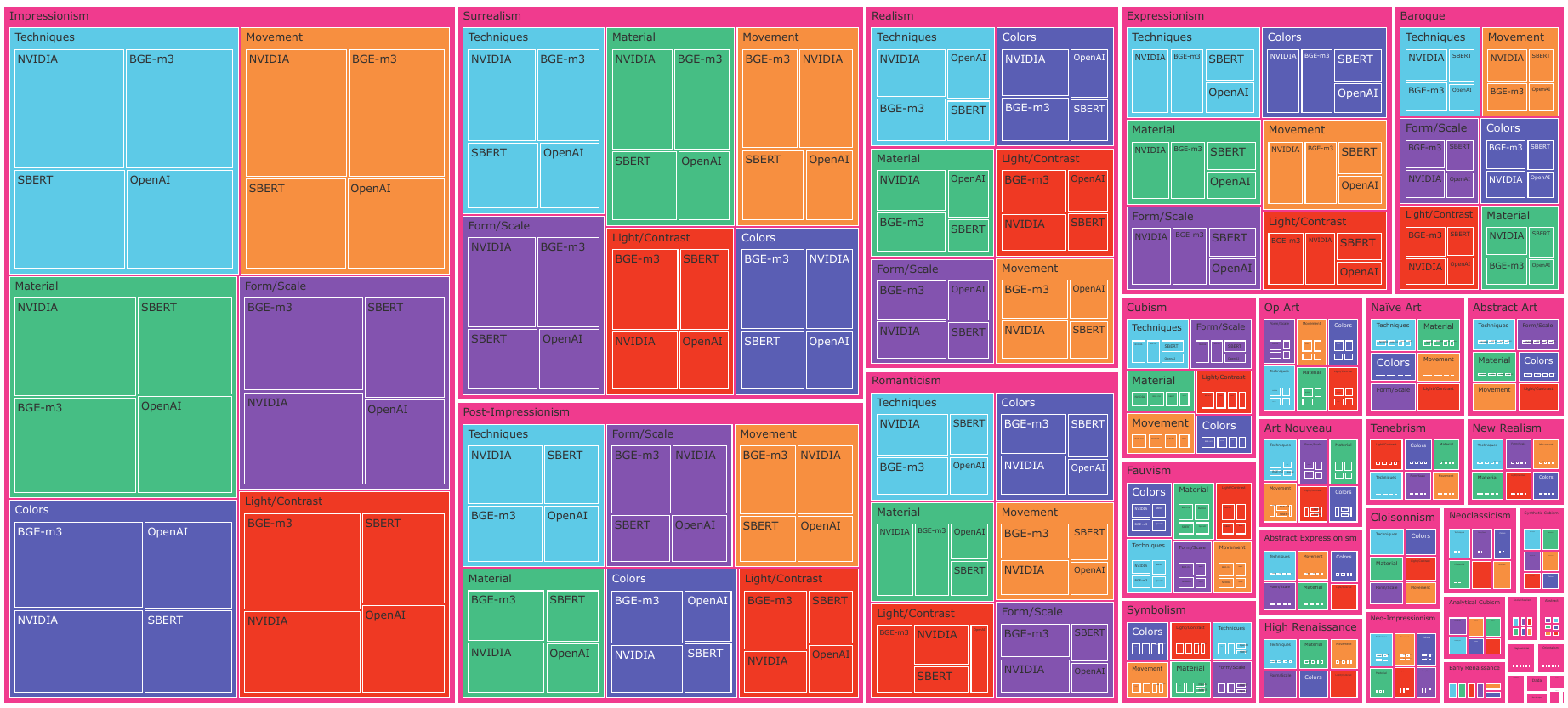}
    \caption{Treemap of the cumulative cosine similarity scores per art style, analysis focus areas and embedding model.}
    \label{fig:fig22S}
\end{figure*}

\begin{figure*}[!ht] 
    \centering
    \includegraphics[angle=90,origin=c,width=\textwidth,height=0.7\textheight,keepaspectratio]{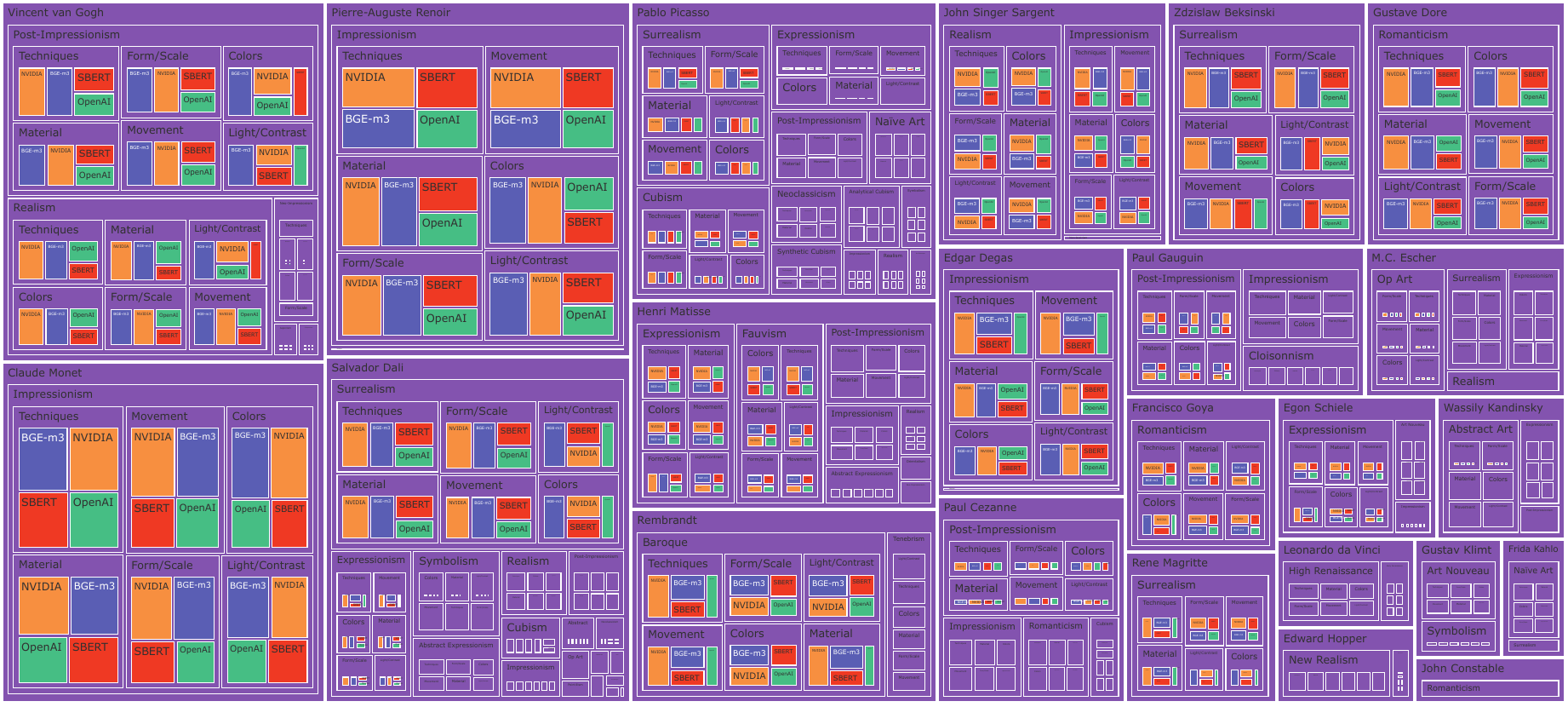}
    \caption{Treemap of the cumulative cosine similarity scores per artist, art style, analysis focus areas and embedding model.}
    \label{fig:fig23S}
\end{figure*}

\end{document}